\pdfoutput=1
\documentclass[11pt]{article}
\usepackage{multicol}
\usepackage{wrapfig}
\usepackage{hyperref}
\usepackage[]{alignment}
\usepackage{times}
\usepackage{latexsym}
\usepackage{arydshln}

\usepackage{placeins}
\usepackage{afterpage}
\usepackage{amsfonts}
\usepackage[T1]{fontenc}
\usepackage[utf8]{inputenc}

\usepackage{microtype}
\usepackage{graphicx}

\usepackage{inconsolata}

\renewcommand\vec{\mathbf}

\newcommand{\treelogo}{\raisebox{5pt}{\includegraphics[scale=0.050]{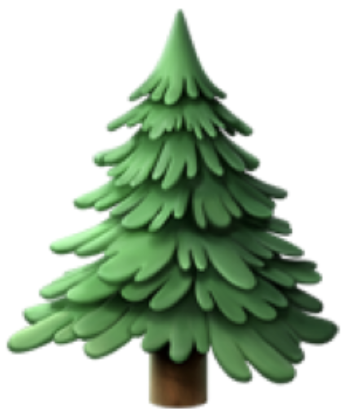}}}
\newcommand{\goog}{\raisebox{3pt}{\includegraphics[trim=600 200 600 0, scale=0.01, clip]{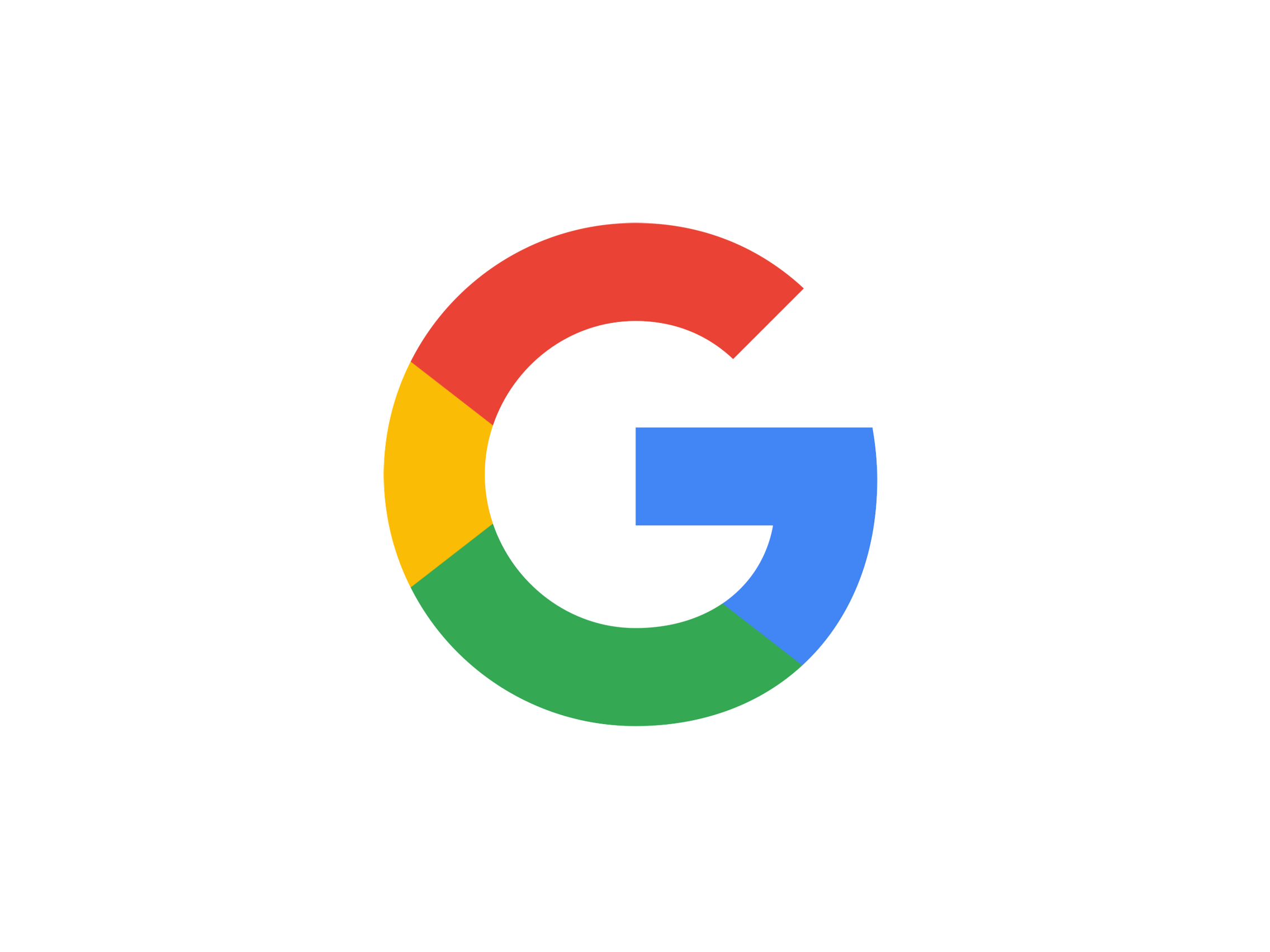}}}
\newcommand{\gtlogo}{\raisebox{3.4pt}{\includegraphics[scale=0.025]{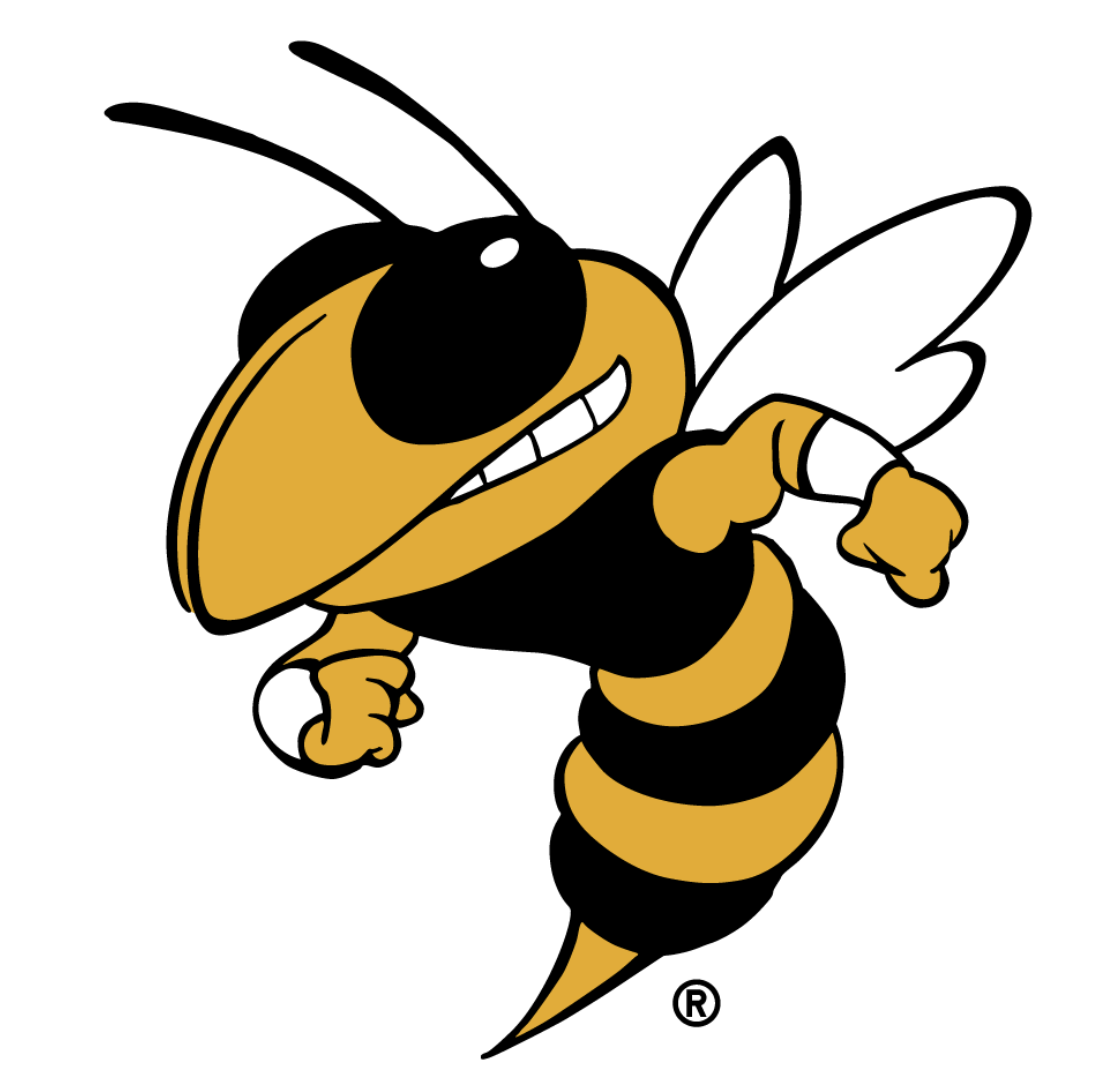}}}

\newcommand{\noog}{\raisebox{5pt}{\includegraphics[scale=0.06]{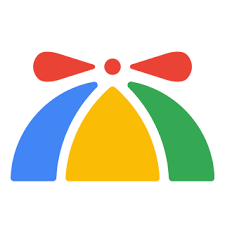}}}
\newcommand\blfootnote[1]{%
  \begingroup
  \renewcommand\thefootnote{}\footnote{#1}%
  \addtocounter{footnote}{-1}%
  \endgroup
}

\title{DAMP: Doubly Aligned Multilingual Parser for Task-Oriented Dialogue}

\author{William Held\noog\gtlogo\hspace{10pt} Christopher Hidey\goog\hspace{10pt}Fei Liu\goog\hspace{10pt}Eric Zhu\goog\\ \textbf{Rahul Goel\goog\hspace{10pt} Diyi Yang\treelogo\hspace{10pt} Rushin Shah\goog}
\\
  \gtlogo Georgia Institute of Technology, \goog Google Assistant, \treelogo Stanford University\\
  \texttt{wheld3@gatech.edu}
  }
\begin{document}
\maketitle

\blfootnote{\noog Work partially done during an internship at Google.}

%Abstract should be concise. Basically, first talk about why this problem is important (1-2 sentences), then what prior work are doing and why they are not enough (1-2 sentences). What we will be doing and how? (1-2 sentences). Lastly, what the results look like compared to prior state of the art models (1-2 sentences)%
\begin{abstract}
Modern virtual assistants use internal semantic parsing engines to convert user utterances to actionable commands. However, prior work has demonstrated multilingual models are less robust for semantic parsing compared to other tasks. In global markets such as India and Latin America, robust multilingual semantic parsing is critical as codeswitching between languages is prevalent for bilingual users. In this work we dramatically improve the zero-shot performance of a multilingual and codeswitched semantic parsing system using two stages of multilingual alignment. First, we show that contrastive alignment pretraining improves \textit{both} English performance and transfer efficiency. We then introduce a constrained optimization approach for hyperparameter-free adversarial alignment during finetuning. Our \textbf{D}oubly \textbf{A}ligned \textbf{M}ultilingual \textbf{P}arser (\textbf{DAMP}) improves mBERT transfer performance by 3x, 6x, and 81x on the Spanglish, Hinglish and Multilingual Task Oriented Parsing benchmarks respectively and outperforms XLM-R and mT5-Large using 3.2x fewer parameters.\footnote{We release code for our constrained optimization technique on \href{https://github.com/SALT-NLP/DAMP}{GitHub}
and finetuned T5 models on \href{https://huggingface.co/models?sort=downloads\&search=willheld\%2Ft5-+mtop}{HuggingFace}.}
\end{abstract}

\section{Introduction}
Task-oriented dialogue systems are the backbone of virtual assistants, an increasingly common direct interaction between users and Natural Language Processing (NLP) technology. Semantic parsing converts unstructured text to structured representations grounded in task actions. Due to the conversational nature of the interaction between users and task-oriented dialogue systems, speakers often use casual register with regional variation. Such variation is an essential challenge for the inclusiveness and reach of virtual assistants which aim to serve a global and diverse userbase~\citep{robustness-eval}. 

In this work, we are motivated by a common form of variation for bilingual speakers~\citep{cs-survey}: codeswitching. Codeswitching occurs in two forms which both affect task-oriented dialogue. Inter-sentential codeswitching is when multilingual users make whole requests in different languages within a single dialogue:

\begin{quote}
    \textbf{Play} all \color[HTML]{ea4335}{rap }\color{black} \textbf{music} on my \color[HTML]{ff9800}iTunes\color{black}\\
    \textbf{Toca} toda la \textbf{música} \color[HTML]{ea4335}{rap }\color{black}en mi \color[HTML]{ff9800}iTunes
\end{quote}

Intra-sentential codeswitching appears when the user switches languages during a single query:

\begin{quote}
   \textbf{Play} toda la \color[HTML]{ea4335}{rap }\color{black}\textbf{music} en mi \color[HTML]{ff9800}iTunes
\end{quote}

Both forms are used by bilingual speakers~\citep{joshi-1982-processing, codeswitch-stats} and cause location, language preference, and even language identification to be unreliable mechanisms for routing requests to an appropriate monolingual system~\citep{cs-lid-challenge}. This makes zero-shot codeswitching performance an aspect of system robustness instead of a way to reduce annotation costs.

However, zero-shot structured prediction and parsing is still a challenge for state-of-the-art multilingual models~\citep{xtremer}, highlighting the need for improved methods beyond scale to achieve this goal. Fortunately, as a fundamental property of the task, these linguistically diverse inputs are grounded in a shared semantic output space. Each of the above outputs corresponds to: 

\begin{quote}
    \textbf{[play\_music:}\color[HTML]{ea4335}{[genre:rap]}\color[HTML]{ff9800}[platform:iTunes]\color{black}\textbf{]}
\end{quote}

This grounded and shared output space makes explicit alignment across languages especially attractive as a mechanism for cross-lingual transfer.

\begin{figure*}[!ht]
\begin{center}
\includegraphics[width=\textwidth]{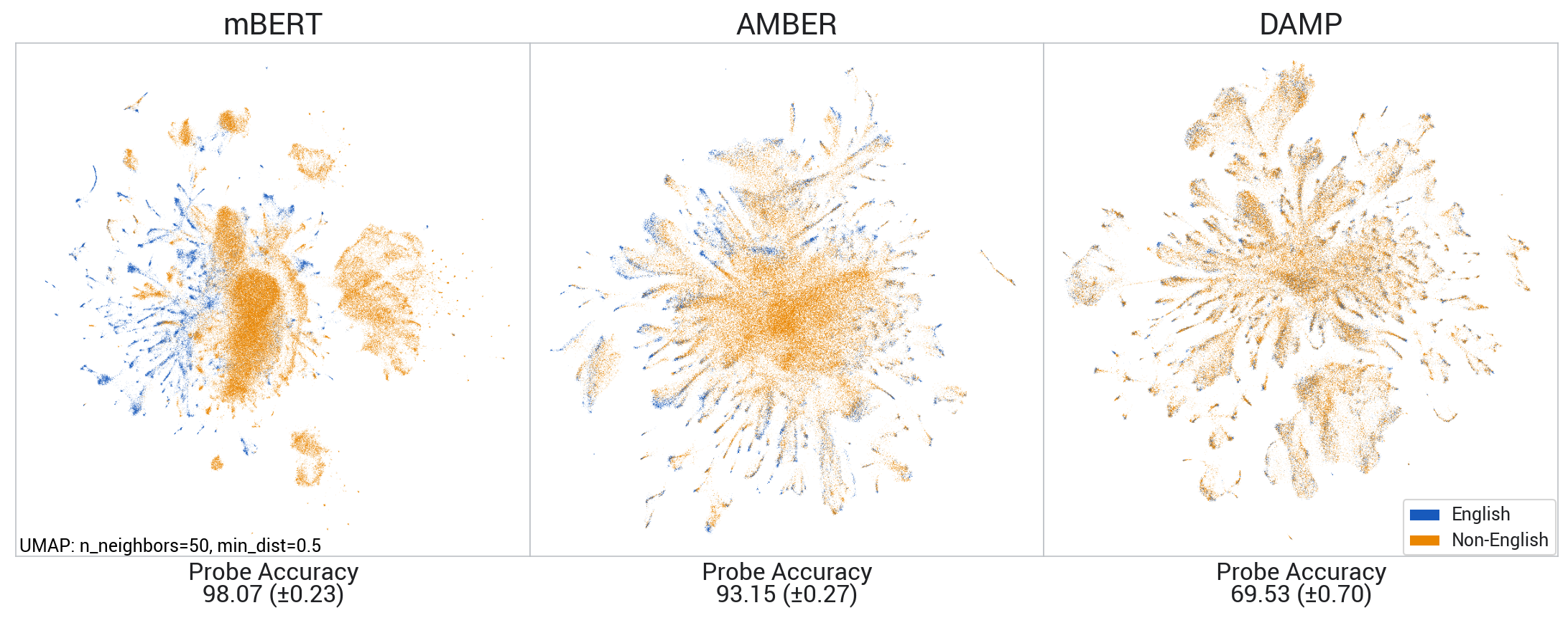}
\caption{We show DAMP meaningfully improves alignment, with more overlapping clusters and decreased probe accuracy. Language identification probe accuracy and visualizations of the token embeddings from a multilingual transformer without alignment (mBERT), pretraining alignment alone (AMBER), and our proposed alignment regime of both contrastive pretraining and constrained adversarial finetuning (DAMP).}\label{embedding_viz}
\end{center}
\end{figure*}

We propose using both contrastive alignment pretraining and a novel constrained adversarial finetuning method to perform \textbf{double alignment}, shown in Figure \ref{embedding_viz}. Our \textbf{D}oubly \textbf{A}ligned \textbf{M}ultilingual \textbf{P}arser (\textbf{DAMP}) achieves strong zero-shot performance on both multilingual (inter-sentential) and intra-sentential codeswitched data, making it a robust model for bilingual users without harming English performance. We contribute the following:
\begin{enumerate}
    \setlength{\itemsep}{0.025in}
    \setlength{\parskip}{0pt}
    \setlength{\parsep}{0pt}
    \item \textbf{Alignment Pretraining Effectiveness:} We show that multilingual BERT (mBERT) has poor transferability for both categories of codeswitched data. Contrastive alignment, however, pretrained with cross-lingual bitext data dramatically improves English, multilingual, and intra-sentential codeswitched semantic parsing performance.

    \item \textbf{Constrained Adversarial Alignment:} We propose utilizing domain adversarial training to further improve alignment and transferability without labeled or aligned data. We introduce a novel constrained optimization method and demonstrate that it improves over prior domain adversarial training algorithms~\citep{prev-adv} and regularization baselines~\citep{norm, freezing}. Finally, we highlight the advantages of pointer-generator networks with explicit alignment by showing that pretrained decoders lead to accidental translation~\citep{mt5-xue}.

    \item \textbf{Interpreting Alignment Improvements:} Additionally, we find the improved parsing ability of DAMP is driven by a 6x improvement in prediction accuracy of the initial intent. Finally, we measure improvements in alignment using a post-hoc linear probe on language prediction in addition to qualitative analysis of embedding visualizations.
\end{enumerate}
\section{Related Work}

\paragraph{Multilingual Language Model Alignment}
Massively multilingual transformers (MMTs)~\citep{mbert, xlmr, mbart, mt5-xue} have become the de-facto basis for multilingual NLP and are effective at intra-sentential codeswitching as well~\citep{mmt-effective-cs}. While prior work has studied explicit alignment of individual embeddings~\citep{artetxe-etal-2018-robust, artetxe-schwenk-2019-massively}, MMTs appear to implicitly perform alignment within their hidden states~\citep{artetxe-etal-2020-cross, conneau-etal-2020-emerging}. 

MMTs are remarkably robust for multilingual and intra-sentential codeswitching benchmarks~\citep{lince, xtreme, xtremer}. However, the gap between performance on the training language and zero-shot targets is larger in task-oriented parsing benchmarks~\citep{mtop, cst5, cstop}, similar to the large discrepancy for other syntactically intensive tasks~\citep{xtreme}.

Our work applies the pretraining regime from \citet{amber}, which adds multiple explicit alignment objectives to traditional MMT pretraining. We show that this technique is effective both for semantic parsing, a new task, and intra-sentential codeswitching, a new linguistic domain.

\paragraph{Domain Adversarial Training}
 The concept of using an adversary to remove undesired features has been discovered and applied separately in transfer learning~\citep{ganin}, privacy preservation~\citep{semi-adv}, and algorithmic fairness~\citep{bias}. When applying this technique to transfer learning, \citet{ganin} term this domain adversarial training. 
 
 Due to its effectiveness in domain transfer learning, multiple works have studied applications of domain adversarial learning to cross-lingual transfer~\citep{adv-detect, adv-extract, adv-gen}. Most relevant, \citet{prev-adv} combine a multi-class language discriminator with translation loss to improve cross-lingual transfer. 
 
In this space, we contribute the 4 following novel findings. Firstly, we show that binary discrimination is more effective than multi-class discrimination and provide intuitive reasoning for why this is true despite the inherently multi-class distribution of multilingual data. Secondly, we show that adversarial alignment can increase the accidental translation phenomena~\citep{mt5-xue} in models with pretrained decoders. Thirdly, we show that token-level adversarial discrimination improves transfer to intra-sentential codeswitching. Finally, we remove the challenge of zero-shot hyperparameter search with a novel constrained optimization technique that can be configured a priori based on our alignment goals.

\paragraph{Preventing Multilingual Forgetting}
Beyond adversarial techniques, prior work has used regularization to maintain multilingual knowledge learned only during pretraining. \citet{norm} shows that penalizing distance from a pretrained model is a simple and effective technique to improve transfer. Using a much stronger inductive bias, \citet{freezing} freezes early layers of multilingual models to preserve multilingual knowledge. This leaves later layers unconstrained for task specific data. We show that DAMP outperforms these baselines, the first comparison of traditional regularization to adversarial cross-lingual transfer.

\section{Methods}
We utilize two separate stages of alignment to improve zero-shot transfer in DAMP. During pretraining, we use contrastive learning to improve alignment amongst pretrained representations. During finetuning, we add \textbf{double} alignment through domain adversarial training using a binary language discriminator and a constrained optimization approach. We apply these improvements to the encoder of a pointer-generator network that copies and generates tags to produce a parse.

\subsection{Baseline Architecture}
Following \citet{pointer-generator}, we use a pointer-generator network to generate semantic parses. We tokenize words $[w_0, w_1\dots, w_m]$ from the labeling scheme into sub-words $[s_{0, w_0},\dots, s_{n, w_0}, s_{0, w_1}\dots, s_{n, w_m}]$ and retrieve hidden states $[\vec{h}_{0, w_0},\dots, \vec{h}_{n, w_0}, \vec{h}_{0, w_1}\dots, \vec{h}_{n, w_m}]$ from our encoder. We use the hidden state of the first subword for each word to produce word-level hidden states: 

\begin{equation}
    [\vec{h}_{0, w_0}, \vec{h}_{0, w_1}\dots, \vec{h}_{0, w_m}]
    \label{hiddens}
\end{equation}

Using \ref{hiddens} as a prefix, we use a randomly initialized auto-regressive decoder to produce representations $[\vec{d}_{0}, \vec{d}_{1}\dots, \vec{d}_{t}]$. At each action-step $a$, we produce a generation logit vector using a perceptron to predict over the vocabulary of intents and slot types $\vec{g}_a$ and a copy logit vector for the arguments from the original query $\vec{c}_a$ using similarity with Eq. \ref{hiddens}:

\begin{equation}
    \vec{g}_a = MLP(\vec{d}_a)
\end{equation}
\begin{equation}
    \vec{c}_a = [\vec{d}_a^\top\vec{h}_{0, w_1}, \vec{d}_a^\top\vec{h}_{0, w_1}, \dots \vec{d}_a^\top\vec{h}_{0, w_m}]
\end{equation}

Finally, we produce a probability distribution $\vec{p}^a$ across both generation and copying by applying the softmax to the concatenation of our logits and optimize the negative log-likelihood of the correct prediction $a'$:

\begin{equation}
\vec{p}^a = \sigma([\vec{g}_a; \vec{c}_a])
\end{equation}
\begin{equation}
    L_s = -log(\vec{p}^a_{a'})
\end{equation}

Intuitively, the pointer-generator limits the model to generating control tokens and copying input tokens. This constraint is key for cross-lingual generalization since our decoder is only trained on English data. Even for models which are pretrained for multilingual generation, finetuning on English data alone often leads to \textit{accidental translation}~\citep{mt5-xue}, where generation occurs in English regardless of the input language.

The pointer-generator guarantees that our generations will use the target language even for languages it was never trained on. We show that this is essential for DAMP in in Section \ref{pre-decoder}, as improved alignment otherwise exacerbates accidental translation by removing the decoders ability to distinguish the input language during generation.

\subsection{Alignment Pretraining}
We evaluate the contrastive pretraining process AMBER introduced by \citet{amber} for semantic parsing. AMBER combines 3 explicit alignment objectives: translation language modeling, sentence alignment, and word alignment using attention symmetry. These procedures aim to make semantically aligned translation data, known as bitext~\citep{melamed-1999-bitext}, similarly aligned in the representation space used by the model.

Translation language modeling was originally proposed by \citet{tlm-paper}. This technique is simply traditional masked language modeling, but uses bitext as input and masking tokens in each language. Since translations of masked words are often unmasked in the bitext, this encourages the model to align word and phrase level representations so that they can be used interchangeably across languages.

Sentence alignment~\citep{xnli} directly optimizes similarity of representations across languages using a siamese network training process. Given an English sentence with pooled representation $\vec{e}_i$, the model maximizes the negative log-likelihood of the probability assigned to true translation $t'$ compared to a batch of possible translations $B$:
\begin{equation}
    L(\vec{e}_i, \vec{t}', N)_{sa} = -\log\left( \frac{\vec{e}_i^\top\vec{t}'}{\sum_{t_i \in B} \vec{e}_i^\top\vec{t}_i}\right)
\end{equation}
 
Finally, AMBER encourages word level alignment by optimizing with an attention symmetry loss~\citep{attend-align}. For attention head $h \in H$, a sentence in language $S$, and its translation in language $T$, the similarity of the cross-attention matrices $A^h_{S\rightarrow T}$ and $A^h_{T\rightarrow S}$ is maximized:
\begin{equation}
    L(S, T) = 1 - \frac{1}{H} \sum_{h \in H} \frac{\mbox{tr}(A^{h\top}_{S\rightarrow T} A^h_{T\rightarrow S})}{\mbox{min}(M, N)}
\end{equation}

Together, these procedures provide signals which encourage the encoder to represent inputs with the same meaning similarly at several levels of granularity, regardless of which language they occur in.

\subsection{Cross-Lingual Adversarial Alignment}
\begin{figure}
\begin{center}
\includegraphics[width=0.48\textwidth]{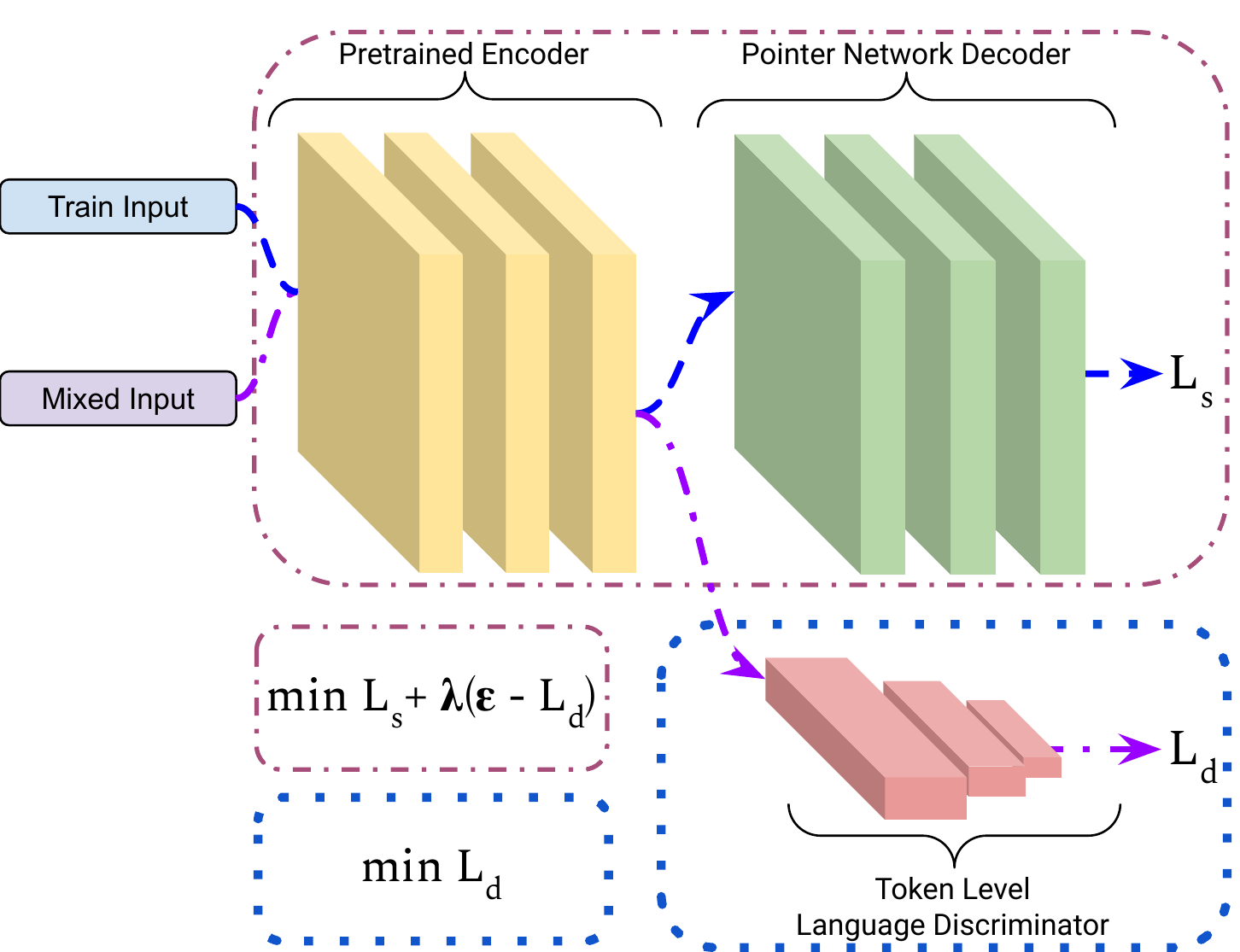}
\caption{An overview of the adversarial alignment procedure. An adversarial model distinguishes English and Non-English examples with $L_d$. With $L_d \geq \epsilon$ as a constraint, the generator optimizes the Lagrangian dual.}\label{model_arch}
\end{center}
\end{figure}
However, this alignment across languages can be lost during finetuning. Since procedures such as those used in AMBER rely on manually aligned data, which is rare for downstream tasks, they are inapplicable for preventing misalignment during finetuning.

Therefore, we instead build on the domain adversarial training process of \citet{ganin} to maintain and improve alignment during finetuning. First, we use a token-level language discriminator as an adversary to maintain word level alignment across languages. We show that multi-class discrimination used in prior work allows for equilibria which are inoptimal for transfer. Instead, we propose treating all languages not found in the training data as a single negative class. Finally, we introduce a general constrained optimization approach for adversarial training and apply it to cross-lingual alignment. 

\paragraph{Token-Level Discriminator}
Similar to \citet{ganin}, we train a discriminator to distinguish between in-domain training data and unlabeled out-of-domain data. Our method assumes access to labeled training queries in one language, in this case English, and unlabeled queries in multiple other languages which target the same intents and slots. Data is sampled evenly from all languages to create an adversarial dataset with equal amounts of each language.

We use a two-layer perceptron to predict the probability $p = P(E|h_{0, w_n})$ that a token with true label $y$ is English or Non-English given hidden representations from Eq. \ref{hiddens}. Our discriminator loss is traditional binary cross-entropy loss:

\begin{equation}
    L_d = -{(y\log(p) + (1 - y)\log(1 - p))}
\end{equation}

Since it is more difficult to discriminate between similar points, domain adversarial training uses the loss of the discriminator as a proxy for alignment. When alignment with the training language improves, so does the cross-lingual transfer to unseen languages.

Prior work using domain adversarial training for multilingual robustness~\citep{adv-extract, prev-adv} performs multi-class classification across all languages and uses the negative log-likelihood of the correct class as the loss function. While using a separate class for each language is natural, it breaks the equivalence between maximizing the discriminator loss and aligning unlabeled and labeled data. With a multi-class discriminator, the generator can instead be rewarded for aligning across unlabeled languages even when this does not benefit transfer from the labeled source.

To illustrate this misaligned reward, suppose we have labeled data in English and unlabeled data in both Spanish and French. The goal of the multi-class adversary is to predict English, Spanish, or French for each token while the encoder is to minimize the ability of the adversary to recover the correct language. Consider the token "dormir", which translates from both Spanish and French to the English "to sleep". In the multi-class setting, the encoder can maximize the adversarial reward by aligning the Spanish "dormir" to the French "dormir", which is simple since they are cognates, without improving alignment with the English "to sleep" at all. In this extreme example, the multi-class loss is likely to lead to a solution which does not improve alignment with the labeled data, in this case English, at all.

Using a binary "English" vs. "Non-English" classifier removes these inoptimal solutions. Since both Spanish and French are now labeled "Non-English", the encoder has no direct incentive to align the two unlabeled languages. Instead, the encoder must align both French \textit{and} Spanish to the labeled English data to the maximize the adversarial reward. Since transferability relies on improved alignment with the labeled data, we expect this loss function to lead to better transfer results.
 
\paragraph{Constrained Optimization}\label{constraint-reasons}
Traditionally, domain adversarial training uses a gradient reversal layer~\citep{ganin} to allow the generator to maximize adversary loss $L_d$ weighted by hyperparameter $\lambda$ while minimizing task loss $L_s$. For the generator, this is effectively equivalent to optimizing a linear combination of the terms:

\begin{equation}
    L = L_s - \lambda L_d
\end{equation}

Selecting a schedule for $\lambda$ presents a challenge in the zero-shot setting. Since the reverse validation procedure used to select the $\lambda$ schedule by \citet{ganin} assumes only one target domain, multilingual works such as \citet{prev-adv} opt to simply perform a linear search using the in-domain development set $s$. This approach ignores transfer performance entirely when weighing adversary loss. Instead, we propose a novel constrained optimization method which balances adversarial and task loss automatically using a constraint derived from first-principles.

Our goal is to obtain token representations that are exactly aligned across languages. Any well-fit adversary will predict English with $P=0.5$ on such data and receives a loss of $0.3$ since it cannot perform better than chance. In equilibrium, the generator cannot increase loss above $0.3$ since the adversary can simply predict $P=0.5$ 
for all inputs regardless of the ground truth labels.

This reasoning provides us a clear constraint. In alignment, the $L_d$ should be no less than $0.3$, which we call $\epsilon$. We then optimize the task loss $L_s$ while enforcing this constraint. We do so with minimal additional computation cost and using back-propagation alone with the differential method of multipliers~\citep{constrained}. The differential method of multipliers first relaxes the constrained problem to its Lagrangian dual: 

\begin{equation}
    L = L_s + \lambda (\epsilon - L_d)
\end{equation}

\begin{figure}
\begin{center}
\includegraphics[trim=0cm 1.2cm 0cm 0.8cm,width=\columnwidth, clip]{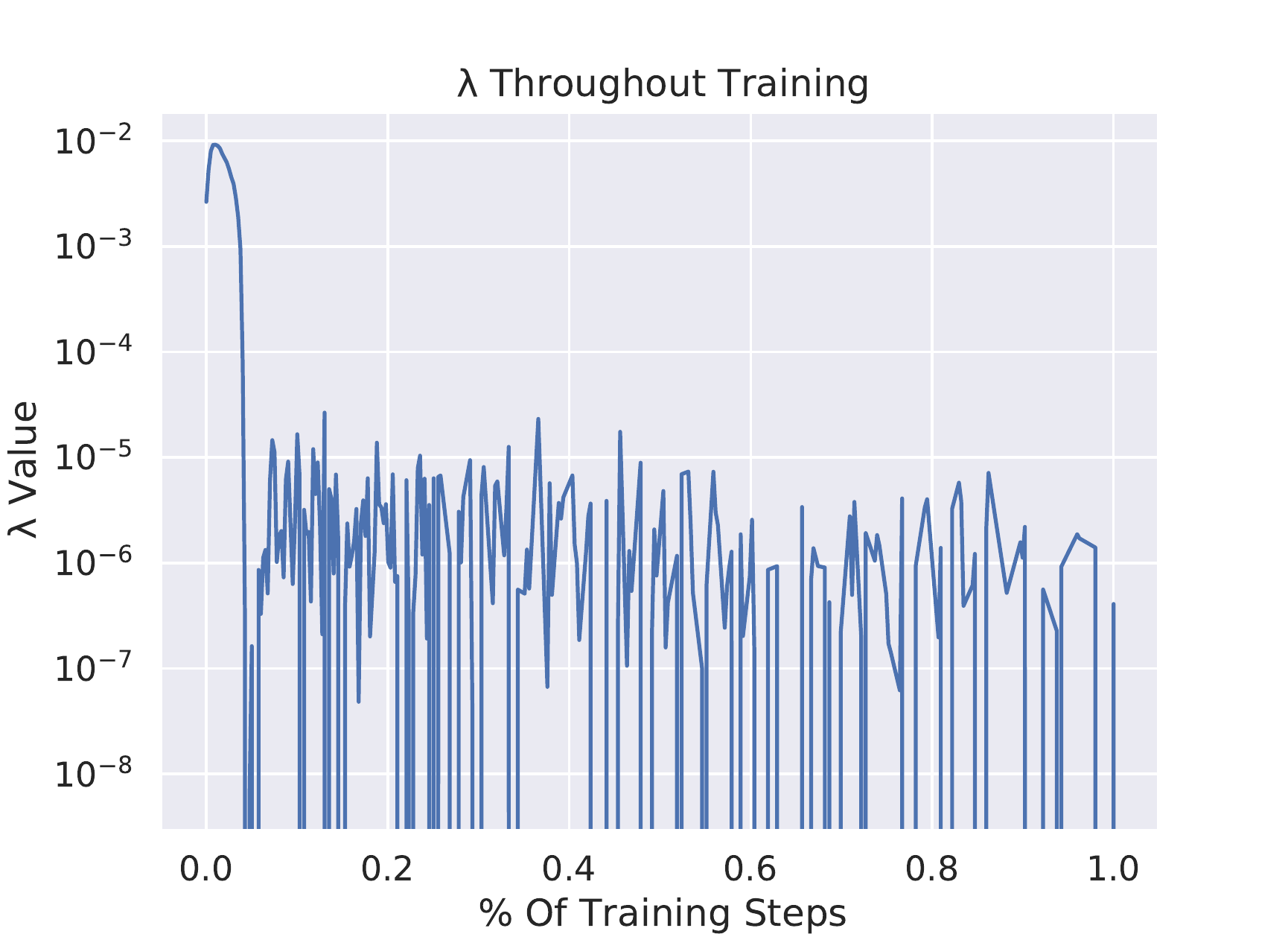}
\includegraphics[trim=0cm 0cm 0cm 0.8cm,width=\columnwidth,clip]{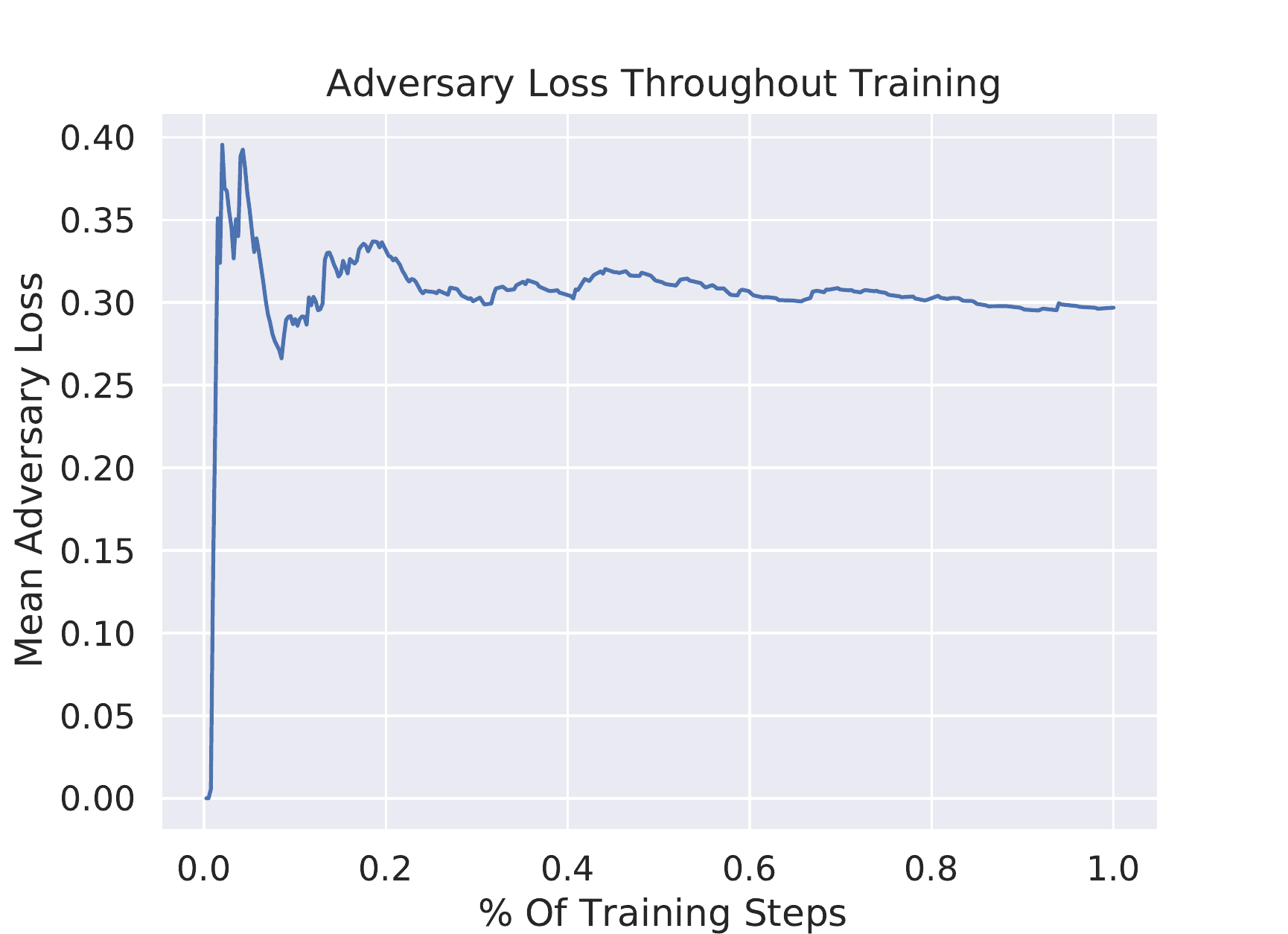}
\caption{The top plot shows the learned schedule for the weight $\lambda$. The bottom plot shows the adversarial loss which converges to our constraint using this $\lambda$ schedule.}\label{constrained}
\end{center}
\end{figure}

Unlike \citet{prev-adv}, this lets us treat $\lambda$ as a learnable parameter and optimize it to maximize the value of $\lambda (\epsilon - L_d)$ with stochastic gradient ascent. In plain terms, our optimization increases the value of $\lambda$ when $\epsilon > L_d$ and decreases it when $\epsilon < L_d$. This produces a schedule for $\lambda$ which weighs the adversarial penalty only when it is accurate. In Figure \ref{constrained}, we show how $\lambda$ evolves throughout training to maintain the constraint.
\section{Experiments}

We evaluate the effects of our techniques on three benchmarks for task-oriented semantic parsing with hierarchical parse structures. Two of these datasets evaluate robustness to intra-sentential codeswitching~\citep{cstop, cst5} and the third uses multilingual data to evaluate robustness to inter-sentential codeswitching~\citep{mtop}. Examples are divided as originally released into training, evaluation, and test data at a ratio of 70/10/20.

\subsection{Datasets}\label{datasets}

\paragraph{Multilingual Task Oriented Parsing (MTOP)}
\citet{mtop} introduced this benchmark to evaluate multilingual transfer for a difficult compositional parse structure. The benchmark contains queries in English, French, Spanish, German, Hindi, and Thai. Zero-shot performance on this benchmark is a proxy for robustness to inter-sentential codeswitching. Each language has approximately 15,000 total queries which cover 11 domains with 117 intents and 78 slot types.

\paragraph{Hindi-English Task Oriented Parsing (CST5)}

\citet{cst5} construct a benchmark of Hindi-English intra-sentential codeswitching data using the same label space as the second version of the English Task Oriented Parsing benchmark~\citep{topv2}. As part of preprocessing, we use \citet{langid} to identify and transliterate Romanized Hindi tokens to Devanagari. There are 125,000 in English and 10,896 queries in Hindi-English which cover 8 domains with 75 Intents and 69 Slot Types.
\begin{table*}
\setlength{\tabcolsep}{5pt}
\begin{center}
\begin{tabular}{l|c|cccccc|cc|}
\cline{2-10}
                                 & en & es & fr & de & hi & th & \multicolumn{1}{c|}{Avg(5 langs)} & Encoder Params. & Ratio \\ \hline
\multicolumn{1}{|l|}{XLM-R$^{*}$}      & 83.9  & 50.3    & 43.9   & \textbf{42.3}   & \textbf{30.9}$^\dagger$  & 26.7 & 38.8                         & 550M       & 3.2x  \\
\multicolumn{1}{|l|}{byT5-Base}  & 80.1 & 13.6 & 11.7 & 10.7 &	1.5	& 2.7	& 8.0 & 436M & 2.5x  \\
\multicolumn{1}{|l|}{mT5-Base}  & 82.5    & 39.0    & 34.9   & 32.6   & 15.7  & 8.3 & 26.1                         & 290M       & 1.7x  \\
\multicolumn{1}{|l|}{mT5-Large$^{**}$}  & 83.2    & 40.0    & 41.1   & 36.2   & 16.5  & 23.0 & 31.4                         & 550M       & 3.2x  \\ \hdashline
\multicolumn{1}{|l|}{mT5-XXL$^{**}$}    & 86.7    & 62.4    & 63.7   & 57.1   & 43.3  & 49.2 & 55.1                         & 6.5B       & 33x \\ \hline
\multicolumn{1}{|l|}{mBERT}      & 78.6    & 0.5     & 1.0    & 0.9    & 0.1   & 0.1  & 0.5                          & 172M       & 1x    \\
\multicolumn{1}{|l|}{AMBER}      & \textbf{84.2}    & 46.4    & 35.8   & 26.3   & 6.7   & 2.7  & 23.6                         & 172M       & 1x    \\
\multicolumn{1}{|l|}{DAMP} & 83.5    & \textbf{56.8}$^\dagger$    & \textbf{55.6}$^\dagger$   & 42.2   & 27.4  & \textbf{29.2}$^\dagger$ & \textbf{42.2}$^\dagger$                         & 172M       & 1x    \\ \hline
\end{tabular}
\caption{Exact Match (EM) accuracy scores on the MTOP dataset. $^{*}$ and $^{**}$ indicate results from \citet{mtop} and \citet{taf-mt5} respectively. Best
results for models which fit on a single consumer GPU in bold. Models marked with $\dagger$ significantly~($p=0.05$) improve over all others using the bootstrap confidence interval.}\label{mtop_main}
\end{center}
\end{table*}
\paragraph{Codeswitching Task Oriented Parsing (CSTOP)}

\citet{cstop} is a benchmark of Spanish-English codeswitching data. While the dataset was released with a corresponding English dataset in the same label space, that data is now unavailable. Therefore, we construct an artificial dataset in the same label space using Google Translate on each segment of the structured Spanish-English training data\footnote{We include the parse brackets during translation to preserve parse structure:  \href{https://cloud.google.com/translate/docs/intro-to-v3\#document}{Google Translate Documents}}. The resulting English dataset is not human validated and therefore noisy. This is a limitation, but is necessary to estimate of zero-shot transfer from English to Spanish-English codeswitching due to the limited release of CSTOP. The resulting dataset has 5,803 queries in both English and Spanish-English which cover 2 domains with 19 Intents and 10 Slot Types.

\subsection{Results}\label{results}
We use the same hyperparameter configurations for all settings. The encoder uses the mBERT architecture~\citep{mbert}. The decoder is a randomly initialized 4-layer, 8-head vanilla transformer for comparison with the 4-layer decoder structure used in \citet{mtop}. We use AdamW and optimize for $1.2$ million training steps with early stopping using a learning rate of $2e{-5}$, batch size of $16$, and decay the learning rate to 0 throughout the training. We train on a Cloud TPU v3 Pod for approximately 4 hours for each dataset. For all adversarial experiments, we use the unlabeled queries from MTOP as training data for our discriminator and a loss constraint $\epsilon$ of $0.3$ as justified in \ref{constraint-reasons}. 

The English data from each benchmark is used for training and early stopping evaluation. We report Exact Match (EM) accuracy on all test splits. In all tables, results that significantly~($p=0.05$) improve over all others are marked with a $\dagger$ using the bootstrap confidence interval~\citep{significance}.

\paragraph{MTOP}

In Table \ref{mtop_main}, we report the results of our training procedure with mBERT, AMBER, and DAMP compared to existing baselines from prior work: XLM-R with a pointer-generator network~\citep{mtop}, MT5~\citep{mt5-xue} and byT5~\cite{xue-etal-2022-byt5}. For both T5 variants, we train with the hyperparameters described in \citet{taf-mt5}.

Despite finetuned mBERT being a strong baseline for other tasks~\citep{freezing, lince, xglue, xtreme, xtremer}, it is ineffective at cross-lingual transfer for compositional semantic parsing achieving an average multilingual accuracy of 0.5.

The AMBER pretraining process significantly improves over mBERT accuracy for all languages to an average of 23.6. Average accuracy across the 5 Non-English languages improves by 47x. English accuracy also improves to 84.2 from 78.6, instead of suffering negative transfer~\citep{negative-interference}.

DAMP further improves average accuracy across languages over AMBER by 1.8x to 42.2, outperforming both similarly sized models (byT5-Base; +34.2, mT5-Base; +16.1) and models three times its size (mT5-Large; +10.8, XLM-R; +3.4). mT5-XXL maintains state-of-the-art performance of 55.1 but requires 33x more parameters and multiple GPUs for inference, which increases latency and compute cost. 

Adversarial alignment improves performance in each language by at least 10 points, with Hindi and Thai, the most distant testing languages from English, having the largest improvements of +20.7 and +26.5 respectively. DAMP improves over the mBERT baseline by 84x without architecture changes or additional inference cost. 

\paragraph{CST5 \& CSTOP}
\begin{table}
\begin{center}
\setlength{\tabcolsep}{5pt}
\begin{tabular}{l|cc|cc|c}
\cline{2-5}
                                & \multicolumn{2}{c|}{CST5}      & \multicolumn{2}{c|}{CSTOP}    & \multicolumn{1}{l}{}       \\ \cline{2-6} 
                                & en            & hi-en          & en            & es-en         & \multicolumn{1}{c|}{Ratio} \\ \hline
\multicolumn{1}{|l|}{byT5-Base} & \textbf{85.5} &	5.5	& 80.0	& 22.3 & \multicolumn{1}{c|}{2.5x} \\
\multicolumn{1}{|l|}{mT5-Base} & \textbf{85.7}             & 14.6            & 80.5             & 28.2             & \multicolumn{1}{c|}{1.7x}  \\
\multicolumn{1}{|l|}{mT5-XXL}   & -             & 20.3         & -             & -             & \multicolumn{1}{c|}{33x}   \\ \hline
\multicolumn{1}{|l|}{mBERT}     & 84.4          & 3.8           & 81.2          & 27.7          & \multicolumn{1}{c|}{1x}    \\
\multicolumn{1}{|l|}{AMBER}     & \textbf{85.8} & 16.7          & \textbf{86.7}$^\dagger$ & 79.3          & \multicolumn{1}{c|}{1x}    \\
\multicolumn{1}{|l|}{DAMP}      & \textbf{85.6}          & \textbf{20.5}$^\dagger$ & 86.0          & \textbf{80.3}$^\dagger$ & \multicolumn{1}{c|}{1x}    \\ \hline
\end{tabular}
\caption{Exact Match (EM) accuracy scores for both intra-sentential codeswitching benchmarks. mT5-XXL results from \citet{cst5}. Best results in bold.}\label{cst5_main}
\end{center}
\end{table}
In Table \ref{cst5_main}, we report the results on both intra-sentential codeswitching benchmarks. For Hindi-English, we compare the MT5-small and MT5-XXL baselines from \citet{cst5}.

AMBER again leads to a performance improvement over mBERT for both CST5 and CSTOP, across English (+1.4, +5.5) and codeswitched (+12.9, +52.4) data. DAMP also further improves transfer results (+3.8, +1.0) over AMBER at the cost of small losses in English performance (-0.2, -0.7). DAMP achieves a new state-of-the-art of 20.5 on zero-shot transfer for CST5, outperforming even MT5-XXL (20.3). Since both alignment stages have word-level objectives, we hypothesize that the word-level inductive bias provides benefits for intra-sentential codeswitching despite lacking explicit supervision for it.

\section{Adversarial Baseline Comparison}
\begin{table}[]
\setlength{\tabcolsep}{2.25pt}
\begin{center}
\begin{tabular}{lcccccc|}
\cline{2-7}
\multicolumn{1}{l|}{}                          & \multicolumn{2}{c|}{MTOP}                   & \multicolumn{2}{c|}{CST5}  & \multicolumn{2}{c|}{CSTOP} \\ \cline{2-7} 
\multicolumn{1}{l|}{}                          & en & \multicolumn{1}{c|}{Avg} & en     & \multicolumn{1}{c|}{hi-en}  & en & es-en   \\ \hline
\multicolumn{7}{|c|}{\textbf{Alignment Ablation}}                                                                         \\ \hline
\multicolumn{1}{|l|}{mBERT}                    & 78.6    & \multicolumn{1}{c|}{0.5}          & 84.4        & \multicolumn{1}{c|}{3.7}     & 81.2 & 27.7     \\
\multicolumn{1}{|l|}{AMBER}                    & \textbf{84.2}    & \multicolumn{1}{c|}{23.6}         & \textbf{85.8}        & \multicolumn{1}{c|}{16.7}      & \textbf{86.7} &    79.3\\
\multicolumn{1}{|l|}{\hspace*{0.25cm}+ Multi}             & 84.0    & \multicolumn{1}{c|}{32.3}         & 85.5        & \multicolumn{1}{c|}{14.1}    & 85.0 & 79.4     \\
\multicolumn{1}{|l|}{\hspace*{0.5cm}+ Constr.} & 82.7    & \multicolumn{1}{c|}{33.7}         & 85.6       & \multicolumn{1}{c|}{13.8}    & 85.1 & \textbf{80.3}     \\
\multicolumn{1}{|l|}{\hspace*{0.25cm}+ Binary}                 & 83.8    & \multicolumn{1}{c|}{35.8}         & \textbf{85.8}        & \multicolumn{1}{c|}{18.4} & 86.3 & 78.1        \\
\multicolumn{1}{|l|}{\hspace*{0.5cm}+ Constr.}     & 83.5    & \multicolumn{1}{c|}{\textbf{42.2}$^\dagger$}         & 85.6        & \multicolumn{1}{c|}{20.5}     & 86.0 & \textbf{80.3}    \\ \hline
\multicolumn{7}{|c|}{\textbf{Regularization Baselines}}                                                                    \\ \hline
\multicolumn{1}{|l|}{\hspace*{0.25cm}+ Freeze}       & 82.6    & \multicolumn{1}{c|}{32.0}         & 85.2        & \multicolumn{1}{c|}{\textbf{24.6}$^\dagger$}      & 85.5 & 77.2   \\
\multicolumn{1}{|l|}{\hspace*{0.25cm}+ $L_{2}$ Norm}      & 81.3    & \multicolumn{1}{c|}{35.5}         & 81.6        & \multicolumn{1}{c|}{22.5}    & 83.4 & 77.5     \\
\multicolumn{1}{|l|}{\hspace*{0.25cm}+ $L_{1}$ Norm}      & 78.6    & \multicolumn{1}{c|}{36.4}         & 80.7        & \multicolumn{1}{c|}{18.7}    & 81.1 & 69.8     \\ \hline
\multicolumn{7}{|c|}{\textbf{Pretrained Decoder Baseline}}                                                                    \\ \hline
\multicolumn{1}{|l|}{\hspace*{0.25cm}mT5-Base}       & 82.5    & \multicolumn{1}{c|}{26.1}         & 85.7        & \multicolumn{1}{c|}{14.6}      & 80.5 & 28.2   \\
\multicolumn{1}{|l|}{\hspace*{0.25cm}+ Align}      & 81.1    & \multicolumn{1}{c|}{16.5}         & 85.5        & \multicolumn{1}{c|}{0.6}    & 83.0 & 16.7     \\
\multicolumn{1}{|l|}{\hspace*{0.25cm}+ Pointer}      & 71.9    & \multicolumn{1}{c|}{15.2}         & 85.0        & \multicolumn{1}{c|}{18.0}    & 77.6 & 54.7     \\ 
\multicolumn{1}{|l|}{\hspace*{0.5cm}+ Align}      & 72.9    & \multicolumn{1}{c|}{20.6}         & 85.0        & \multicolumn{1}{c|}{3.6}    & 80.6 & 56.1     \\ \hline
\end{tabular}
\caption{Exact Match (EM) accuracy scores for across combinations of both binary and multi-class discriminators, constrained optimization, and regularization.}\label{tablation}
\end{center}
\end{table}

\subsection{Adversary Ablation} In Table \ref{tablation}, we isolate the effects of our contributions to domain adversarial training with an ablation study. While all adversarial variants improve transfer results, we see that using a binary adversary and our constrained optimization technique are both mutually and independently beneficial to adversarial alignment. Notably, DAMP improves over the unconstrained multi-class adversarial technique used in \citet{prev-adv} by 9.9, 6.4, and 0.9 EM accuracy points on MTOP, CST5, and CSTOP respectively.

\subsection{Regularization Comparison} We also compare adversarial training to regularization techniques used in cross-lingual learning. We experiment with freezing the first 8 layers of the encoder~\citep{freezing} and using the $L_1$ and $L_2$ norm penalty~\citep{norm}. Adversarial learning outperforms these baselines on MTOP and CSTOP while model freezing and $L_2$ norm penalization outperform adversarial learning on CST5. However, adversarial learning is the only method that improves across all benchmarks.

\subsection{Pretrained Decoder Comparison}\label{pre-decoder} Finally, we evaluate whether our constrained adversarial alignment technique offers similar benefits to models with pretrained decoders due to their natural advantage in generation tasks. We find that adversarial training does worse than the plain mT5 model (-9.6). Upon inspection, adversarial alignment causes this drop by exacerbating \textit{accidental translation}~\citep{mt5-xue}, where the output for Non-English input is translated to English.

For example, the expected output for ``Merci d’envoyer la ligne de travail`` is ``[IN:SEND\_MESSAGE [SL:GROUP travail]]``. While the unaligned model produces the incorrect parse ``[IN:SEND\_MESSAGE [SL:RECIPIENT la ligne de travail]]``, the aligned model produces the correct parse translated to English ``[IN:SEND\_MESSAGE [SL:GROUP work]]``. In DAMP, the pointer-generator fundamentally prevents accidental translation.

We confirm this in mT5 by reformatting the decoding task in a pointer format, where the correct output in the above example would be ``[IN:SEND\_MESSAGE [SL:GROUP <pt-5>]]``. This makes accidental translation impossible, and adversarial alignment again improves performance in this variant for MTOP and CSTOP. However, the mT5 decoder struggles to adapt to this task, making overall performance worse than DAMP.
\subsection{Improvement Analysis}
\begin{table}[]
\begin{center}
\setlength{\tabcolsep}{1.25pt}
\begin{tabular}{l|c|cccccr|}
\cline{2-8}
 & en & es & fr & de & hi & th & \multicolumn{1}{l|}{Avg} \\ \hline
\multicolumn{1}{|l|}{mBERT} & 94.7 & 15.3 & 17.0 & 10.7 & 7.0 & 8.2 & 11.6 \\
\multicolumn{1}{|l|}{AMBER} & \textbf{96.4} & 78.7 & 71.3 & 66.3 & 32.5 & 26.5 & 55.1 \\
\multicolumn{1}{|l|}{DAMP} & \textbf{96.4} & \textbf{89.0}$^\dagger$ & \textbf{86.4}$^\dagger$ & \textbf{80.5}$^\dagger$ & \textbf{76.6}$^\dagger$ & \textbf{74.4}$^\dagger$ & \textbf{81.4}$^\dagger$ \\ \hline
\end{tabular}
\caption{Intent Prediction accuracy for each language on the MTOP dataset for mBERT, AMBER, and DAMP.}\label{ip_improvements}
\end{center}
\end{table}
Since exact match accuracy is a strict metric, we analyze our improvements with qualitative analysis. We examine examples that DAMP predicts correctly but AMBER and mBERT do not. We then randomly sample 20 examples from each language for manual evaluation. 

Improvements in intent prediction are a large portion of the gain. If intent prediction fails, the rest of the auto-regressive decoding goes awry as the decoder attempts to generate valid slot types for that intent. We report intent prediction results across the test dataset in Table \ref{ip_improvements}.

In general, these improvements follow a trend from nonsensical errors to reasonable errors to correct. For example, given the French phrase ``S’il te plait appelle Adam.'' meaning ``Please call Adam."", mBERT predicts the intent \textit{QUESTION\_MUSIC}, AMBER predicts \textit{GET\_INFO\_CONTACT}, and DAMP predicts the correct \textit{CREATE\_CALL}.

Within the slots themselves, the primary improvements noted in DAMP are more accurate placement articles and prepositions such as "du", "a", "el", and "la" inside the slot boundaries, which is of arguable real world importance. 

We present the full sample of examples used for this analysis in Tables \ref{first-qual}-\ref{final-qual} in the Appendix.

\section{Alignment Analysis}
We analyze how well our alignment goals are met using two methods in Figure \ref{embedding_viz}. First, we use a two-dimensional projection of the resulting encoder embeddings to provide a visual intuition for alignment. Then, we provide a more reliable quantitatively evaluate alignment using a post-hoc linear probe.

\subsection{Embedding Space Visualization}
 In Figure \ref{embedding_viz}, we visualize the embedding spaces of each model variant on each MTOP test set using Universal Manifold Approximation and Projection (UMAP)~\citep{umap}. Our visualization of mBERT provides a strong intuition for its poor results, as English and Non-English data form linearly separate clusters even within this reduced embedding space. By using AMBER instead, this global clustering behavior is removed and replaced by small local clusters of English and Non-English data. Finally, DAMP produces an embedding space with no clear visual clusters of Non-English data without English data intermingled.

\subsection{Post-Hoc Probing}
We evaluate improvements to alignment quantitatively. While \citet{prev-adv} reports the performance of the training adversary as evidence of successful training, this method has been shown insufficient due to mode collapse during training~\citep{dem-remove, rlace}. Therefore, we train a linear probe on a frozen model after training for each variant using 10-fold cross-validation. 

Supporting the visual intuition, probe performance decreases with each stage of alignment. On mBERT, the discriminator achieves 98.07 percent accuracy indicating poor alignment. AMBER helps, but the discriminator still achieves 93.15 percent accuracy indicating the need for further removal. DAMP results in a 23.62 point drop in discriminator accuracy to 69.53. This is still far above chance despite our training adversary converging to close-to-random accuracy. This indicates both the need for post-hoc probing and the possibility of further alignment improvements.

\section{Conclusions}
In this work, we introduce a Doubly Aligned Multilingual Parser (DAMP), a semantic parsing training regime that uses contrastive alignment pretraining and adversarial alignment during fine-tuning with a novel constrained optimization approach. 
We demonstrate that both of these stages of alignment benefit transfer learning in semantic parsing to both inter-sentential (multilingual) and intra-sentential codemixed data, outperforming both similarly sized and larger models. We analyze the effects of DAMP, comparing our proposed alignment method broadly to prior both adversarial techniques and regularization baselines, and its generalizability, with applications to pretrained decoders. Finally, we interpret the impacts of both stages of alignment through qualitative improvement analysis and quantitative probing. 

Importantly, DAMP shows that alignment in \textit{both} pretraining and finetuning can outperform larger models pretrained on more data. This offers an orthogonal improvement to the current scaling paradigm, supporting the idea that current multilingual models underutilize available bitext~\citep{reid2022role}. In cases where bitext is unavailable, our work shows that alignment still possible via adversarial procedures. By releasing our simplified constrained optimization approach for multilingual adversarial alignment, we aim to simplify and improve the application of such approaches for future work.

\section{Limitations}\label{limitation}
This work only carries out experiments using English as the base training language for domain adversarial transfer. It is possible that domain adversarial transfer has a variable effect depending on the training language from which labeled data is used. Additionally, while typologically and regionally diverse, all but one language used in our evaluation is of Indo-European origin.

\section{Acknowledgements}
We are thankful to Hongxin Zhang, Caleb Ziems, and the anonymous reviewers from Google, ACL Rolling Review, and the ACL Main Conference for their helpful feedback.

\bibliography{custom}

\begin{thebibliography}{47}
\expandafter\ifx\csname natexlab\endcsname\relax\def\natexlab#1{#1}\fi

\bibitem[{Agarwal et~al.(2022)Agarwal, Gupta, Goel, Upadhyay, Joshi, and
  Aravamudhan}]{cst5}
Anmol Agarwal, Jigar Gupta, Rahul Goel, Shyam Upadhyay, Pankaj Joshi, and
  Rengarajan Aravamudhan. 2022.
\newblock Cst5: Data augmentation for code-switched semantic parsing.
\newblock \emph{arXiv preprint arXiv:2211.07514}.

\bibitem[{Aguilar et~al.(2020)Aguilar, Kar, and Solorio}]{lince}
Gustavo Aguilar, Sudipta Kar, and Thamar Solorio. 2020.
\newblock \href {https://www.aclweb.org/anthology/2020.lrec-1.223} {{L}in{CE}:
  {A} {C}entralized {B}enchmark for {L}inguistic {C}ode-switching
  {E}valuation}.
\newblock In \emph{Proceedings of The 12th Language Resources and Evaluation
  Conference}, pages 1803--1813, Marseille, France. European Language Resources
  Association.

\bibitem[{Artetxe et~al.(2018)Artetxe, Labaka, and
  Agirre}]{artetxe-etal-2018-robust}
Mikel Artetxe, Gorka Labaka, and Eneko Agirre. 2018.
\newblock \href {https://doi.org/10.18653/v1/P18-1073} {A robust self-learning
  method for fully unsupervised cross-lingual mappings of word embeddings}.
\newblock In \emph{Proceedings of the 56th Annual Meeting of the Association
  for Computational Linguistics (Volume 1: Long Papers)}, pages 789--798,
  Melbourne, Australia. Association for Computational Linguistics.

\bibitem[{Artetxe et~al.(2020)Artetxe, Ruder, and
  Yogatama}]{artetxe-etal-2020-cross}
Mikel Artetxe, Sebastian Ruder, and Dani Yogatama. 2020.
\newblock \href {https://doi.org/10.18653/v1/2020.acl-main.421} {On the
  cross-lingual transferability of monolingual representations}.
\newblock In \emph{Proceedings of the 58th Annual Meeting of the Association
  for Computational Linguistics}, pages 4623--4637, Online. Association for
  Computational Linguistics.

\bibitem[{Artetxe and Schwenk(2019)}]{artetxe-schwenk-2019-massively}
Mikel Artetxe and Holger Schwenk. 2019.
\newblock \href {https://doi.org/10.1162/tacl_a_00288} {Massively multilingual
  sentence embeddings for zero-shot cross-lingual transfer and beyond}.
\newblock \emph{Transactions of the Association for Computational Linguistics},
  7:597--610.

\bibitem[{Barman et~al.(2014)Barman, Das, Wagner, and
  Foster}]{cs-lid-challenge}
Utsab Barman, Amitava Das, Joachim Wagner, and Jennifer Foster. 2014.
\newblock \href {https://doi.org/10.3115/v1/W14-3902} {Code mixing: A challenge
  for language identification in the language of social media}.
\newblock In \emph{Proceedings of the First Workshop on Computational
  Approaches to Code Switching}, pages 13--23, Doha, Qatar. Association for
  Computational Linguistics.

\bibitem[{Chen et~al.(2020)Chen, Ghoshal, Mehdad, Zettlemoyer, and
  Gupta}]{topv2}
Xilun Chen, Asish Ghoshal, Yashar Mehdad, Luke Zettlemoyer, and Sonal Gupta.
  2020.
\newblock \href {https://doi.org/10.18653/v1/2020.emnlp-main.413} {Low-resource
  domain adaptation for compositional task-oriented semantic parsing}.
\newblock In \emph{Proceedings of the 2020 Conference on Empirical Methods in
  Natural Language Processing (EMNLP)}, pages 5090--5100, Online. Association
  for Computational Linguistics.

\bibitem[{Cohn et~al.(2016)Cohn, Hoang, Vymolova, Yao, Dyer, and
  Haffari}]{attend-align}
Trevor Cohn, Cong Duy~Vu Hoang, Ekaterina Vymolova, Kaisheng Yao, Chris Dyer,
  and Gholamreza Haffari. 2016.
\newblock \href {https://doi.org/10.18653/v1/N16-1102} {Incorporating
  structural alignment biases into an attentional neural translation model}.
\newblock In \emph{Proceedings of the 2016 Conference of the North {A}merican
  Chapter of the Association for Computational Linguistics: Human Language
  Technologies}, pages 876--885, San Diego, California. Association for
  Computational Linguistics.

\bibitem[{Conneau et~al.(2020{\natexlab{a}})Conneau, Khandelwal, Goyal,
  Chaudhary, Wenzek, Guzm{\'a}n, Grave, Ott, Zettlemoyer, and Stoyanov}]{xlmr}
Alexis Conneau, Kartikay Khandelwal, Naman Goyal, Vishrav Chaudhary, Guillaume
  Wenzek, Francisco Guzm{\'a}n, Edouard Grave, Myle Ott, Luke Zettlemoyer, and
  Veselin Stoyanov. 2020{\natexlab{a}}.
\newblock \href {https://doi.org/10.18653/v1/2020.acl-main.747} {Unsupervised
  cross-lingual representation learning at scale}.
\newblock In \emph{Proceedings of the 58th Annual Meeting of the Association
  for Computational Linguistics}, pages 8440--8451, Online. Association for
  Computational Linguistics.

\bibitem[{Conneau and Lample(2019)}]{tlm-paper}
Alexis Conneau and Guillaume Lample. 2019.
\newblock \href
  {https://proceedings.neurips.cc/paper/2019/file/c04c19c2c2474dbf5f7ac4372c5b9af1-Paper.pdf}
  {Cross-lingual language model pretraining}.
\newblock In \emph{Advances in Neural Information Processing Systems},
  volume~32. Curran Associates, Inc.

\bibitem[{Conneau et~al.(2018)Conneau, Rinott, Lample, Williams, Bowman,
  Schwenk, and Stoyanov}]{xnli}
Alexis Conneau, Ruty Rinott, Guillaume Lample, Adina Williams, Samuel Bowman,
  Holger Schwenk, and Veselin Stoyanov. 2018.
\newblock \href {https://doi.org/10.18653/v1/D18-1269} {{XNLI}: Evaluating
  cross-lingual sentence representations}.
\newblock In \emph{Proceedings of the 2018 Conference on Empirical Methods in
  Natural Language Processing}, pages 2475--2485, Brussels, Belgium.
  Association for Computational Linguistics.

\bibitem[{Conneau et~al.(2020{\natexlab{b}})Conneau, Wu, Li, Zettlemoyer, and
  Stoyanov}]{conneau-etal-2020-emerging}
Alexis Conneau, Shijie Wu, Haoran Li, Luke Zettlemoyer, and Veselin Stoyanov.
  2020{\natexlab{b}}.
\newblock \href {https://doi.org/10.18653/v1/2020.acl-main.536} {Emerging
  cross-lingual structure in pretrained language models}.
\newblock In \emph{Proceedings of the 58th Annual Meeting of the Association
  for Computational Linguistics}, pages 6022--6034, Online. Association for
  Computational Linguistics.

\bibitem[{Dey and Fung(2014)}]{codeswitch-stats}
Anik Dey and Pascale Fung. 2014.
\newblock \href
  {http://www.lrec-conf.org/proceedings/lrec2014/pdf/922_Paper.pdf} {A
  {H}indi-{E}nglish code-switching corpus}.
\newblock In \emph{Proceedings of the Ninth International Conference on
  Language Resources and Evaluation ({LREC}'14)}, Reykjavik, Iceland. European
  Language Resources Association (ELRA).

\bibitem[{Do{\u{g}}ru{\"o}z et~al.(2021)Do{\u{g}}ru{\"o}z, Sitaram, Bullock,
  and Toribio}]{cs-survey}
A.~Seza Do{\u{g}}ru{\"o}z, Sunayana Sitaram, Barbara~E. Bullock, and
  Almeida~Jacqueline Toribio. 2021.
\newblock \href {https://doi.org/10.18653/v1/2021.acl-long.131} {A survey of
  code-switching: Linguistic and social perspectives for language
  technologies}.
\newblock In \emph{Proceedings of the 59th Annual Meeting of the Association
  for Computational Linguistics and the 11th International Joint Conference on
  Natural Language Processing (Volume 1: Long Papers)}, pages 1654--1666,
  Online. Association for Computational Linguistics.

\bibitem[{Dror et~al.(2018)Dror, Baumer, Shlomov, and Reichart}]{significance}
Rotem Dror, Gili Baumer, Segev Shlomov, and Roi Reichart. 2018.
\newblock \href {https://doi.org/10.18653/v1/P18-1128} {The hitchhiker{'}s
  guide to testing statistical significance in natural language processing}.
\newblock In \emph{Proceedings of the 56th Annual Meeting of the Association
  for Computational Linguistics (Volume 1: Long Papers)}, pages 1383--1392,
  Melbourne, Australia. Association for Computational Linguistics.

\bibitem[{Einolghozati et~al.(2021)Einolghozati, Arora, Sainz-Maza~Lecanda,
  Kumar, and Gupta}]{cstop}
Arash Einolghozati, Abhinav Arora, Lorena Sainz-Maza~Lecanda, Anuj Kumar, and
  Sonal Gupta. 2021.
\newblock \href {https://doi.org/10.18653/v1/2021.eacl-main.87} {El volumen
  louder por favor: Code-switching in task-oriented semantic parsing}.
\newblock In \emph{Proceedings of the 16th Conference of the European Chapter
  of the Association for Computational Linguistics: Main Volume}, pages
  1009--1021, Online. Association for Computational Linguistics.

\bibitem[{Elazar and Goldberg(2018)}]{dem-remove}
Yanai Elazar and Yoav Goldberg. 2018.
\newblock \href {https://doi.org/10.18653/v1/D18-1002} {Adversarial removal of
  demographic attributes from text data}.
\newblock In \emph{Proceedings of the 2018 Conference on Empirical Methods in
  Natural Language Processing}, pages 11--21, Brussels, Belgium. Association
  for Computational Linguistics.

\bibitem[{Ganin et~al.(2016)Ganin, Ustinova, Ajakan, Germain, Larochelle,
  Laviolette, Marchand, and Lempitsky}]{ganin}
Yaroslav Ganin, Evgeniya Ustinova, Hana Ajakan, Pascal Germain, Hugo
  Larochelle, Fran\c{c}ois Laviolette, Mario Marchand, and Victor Lempitsky.
  2016.
\newblock Domain-adversarial training of neural networks.
\newblock \emph{J. Mach. Learn. Res.}, 17(1):2096–2030.

\bibitem[{Guzman-Nateras et~al.(2022)Guzman-Nateras, Nguyen, and
  Nguyen}]{adv-detect}
Luis Guzman-Nateras, Minh~Van Nguyen, and Thien Nguyen. 2022.
\newblock \href {https://doi.org/10.18653/v1/2022.naacl-main.409}
  {Cross-lingual event detection via optimized adversarial training}.
\newblock In \emph{Proceedings of the 2022 Conference of the North American
  Chapter of the Association for Computational Linguistics: Human Language
  Technologies}, pages 5588--5599, Seattle, United States. Association for
  Computational Linguistics.

\bibitem[{Hu et~al.(2021)Hu, Johnson, Firat, Siddhant, and Neubig}]{amber}
Junjie Hu, Melvin Johnson, Orhan Firat, Aditya Siddhant, and Graham Neubig.
  2021.
\newblock \href {https://doi.org/10.18653/v1/2021.naacl-main.284} {Explicit
  alignment objectives for multilingual bidirectional encoders}.
\newblock In \emph{Proceedings of the 2021 Conference of the North American
  Chapter of the Association for Computational Linguistics: Human Language
  Technologies}, pages 3633--3643, Online. Association for Computational
  Linguistics.

\bibitem[{Hu et~al.(2020)Hu, Ruder, Siddhant, Neubig, Firat, and
  Johnson}]{xtreme}
Junjie Hu, Sebastian Ruder, Aditya Siddhant, Graham Neubig, Orhan Firat, and
  Melvin Johnson. 2020.
\newblock \href {http://proceedings.mlr.press/v119/hu20b.html} {Xtreme: A
  massively multilingual multi-task benchmark for evaluating cross-lingual
  generalisation}.
\newblock In \emph{ICML}, pages 4411--4421.

\bibitem[{Joshi(1982)}]{joshi-1982-processing}
Aravind~K. Joshi. 1982.
\newblock \href {https://aclanthology.org/C82-1023} {Processing of sentences
  with intra-sentential code-switching}.
\newblock In \emph{{C}oling 1982: Proceedings of the {N}inth {I}nternational
  {C}onference on {C}omputational {L}inguistics}.

\bibitem[{Joty et~al.(2017)Joty, Nakov, M{\`a}rquez, and Jaradat}]{adv-gen}
Shafiq Joty, Preslav Nakov, Llu{\'\i}s M{\`a}rquez, and Israa Jaradat. 2017.
\newblock \href {https://doi.org/10.18653/v1/K17-1024} {Cross-language learning
  with adversarial neural networks}.
\newblock In \emph{Proceedings of the 21st Conference on Computational Natural
  Language Learning ({C}o{NLL} 2017)}, pages 226--237, Vancouver, Canada.
  Association for Computational Linguistics.

\bibitem[{Lange et~al.(2020)Lange, Iurshina, Adel, and
  Str{\"o}tgen}]{adv-extract}
Lukas Lange, Anastasiia Iurshina, Heike Adel, and Jannik Str{\"o}tgen. 2020.
\newblock \href {https://doi.org/10.18653/v1/2020.repl4nlp-1.14} {Adversarial
  alignment of multilingual models for extracting temporal expressions from
  text}.
\newblock In \emph{Proceedings of the 5th Workshop on Representation Learning
  for NLP}, pages 103--109, Online. Association for Computational Linguistics.

\bibitem[{Li et~al.(2021)Li, Arora, Chen, Gupta, Gupta, and Mehdad}]{mtop}
Haoran Li, Abhinav Arora, Shuohui Chen, Anchit Gupta, Sonal Gupta, and Yashar
  Mehdad. 2021.
\newblock \href {https://doi.org/10.18653/v1/2021.eacl-main.257} {{MTOP}: A
  comprehensive multilingual task-oriented semantic parsing benchmark}.
\newblock In \emph{Proceedings of the 16th Conference of the European Chapter
  of the Association for Computational Linguistics: Main Volume}, pages
  2950--2962, Online. Association for Computational Linguistics.

\bibitem[{Li et~al.(2018)Li, Grandvalet, and Davoine}]{norm}
Xuhong Li, Yves Grandvalet, and Franck Davoine. 2018.
\newblock \href {https://proceedings.mlr.press/v80/li18a.html} {Explicit
  inductive bias for transfer learning with convolutional networks}.
\newblock In \emph{Proceedings of the 35th International Conference on Machine
  Learning}, volume~80 of \emph{Proceedings of Machine Learning Research},
  pages 2825--2834. PMLR.

\bibitem[{Liang et~al.(2020)Liang, Duan, Gong, Wu, Guo, Qi, Gong, Shou, Jiang,
  Cao, Fan, Zhang, Agrawal, Cui, Wei, Bharti, Qiao, Chen, Wu, Liu, Yang,
  Campos, Majumder, and Zhou}]{xglue}
Yaobo Liang, Nan Duan, Yeyun Gong, Ning Wu, Fenfei Guo, Weizhen Qi, Ming Gong,
  Linjun Shou, Daxin Jiang, Guihong Cao, Xiaodong Fan, Ruofei Zhang, Rahul
  Agrawal, Edward Cui, Sining Wei, Taroon Bharti, Ying Qiao, Jiun-Hung Chen,
  Winnie Wu, Shuguang Liu, Fan Yang, Daniel Campos, Rangan Majumder, and Ming
  Zhou. 2020.
\newblock \href {https://doi.org/10.18653/v1/2020.emnlp-main.484} {{XGLUE}: A
  new benchmark datasetfor cross-lingual pre-training, understanding and
  generation}.
\newblock In \emph{Proceedings of the 2020 Conference on Empirical Methods in
  Natural Language Processing (EMNLP)}, pages 6008--6018, Online. Association
  for Computational Linguistics.

\bibitem[{Liu et~al.(2021)Liu, Takanobu, Wen, Wan, Li, Nie, Li, Peng, and
  Huang}]{robustness-eval}
Jiexi Liu, Ryuichi Takanobu, Jiaxin Wen, Dazhen Wan, Hongguang Li, Weiran Nie,
  Cheng Li, Wei Peng, and Minlie Huang. 2021.
\newblock \href {https://doi.org/10.18653/v1/2021.acl-long.192} {Robustness
  testing of language understanding in task-oriented dialog}.
\newblock In \emph{Proceedings of the 59th Annual Meeting of the Association
  for Computational Linguistics and the 11th International Joint Conference on
  Natural Language Processing (Volume 1: Long Papers)}, pages 2467--2480,
  Online. Association for Computational Linguistics.

\bibitem[{Liu et~al.(2020)Liu, Gu, Goyal, Li, Edunov, Ghazvininejad, Lewis, and
  Zettlemoyer}]{mbart}
Yinhan Liu, Jiatao Gu, Naman Goyal, Xian Li, Sergey Edunov, Marjan
  Ghazvininejad, Mike Lewis, and Luke Zettlemoyer. 2020.
\newblock \href {https://doi.org/10.1162/tacl_a_00343} {Multilingual denoising
  pre-training for neural machine translation}.
\newblock \emph{Transactions of the Association for Computational Linguistics},
  8:726--742.

\bibitem[{McInnes et~al.(2018)McInnes, Healy, Saul, and Gro{\ss}berger}]{umap}
Leland McInnes, John Healy, Nathaniel Saul, and Lukas Gro{\ss}berger. 2018.
\newblock Umap: Uniform manifold approximation and projection.
\newblock \emph{Journal of Open Source Software}, 3(29):861.

\bibitem[{Melamed(1999)}]{melamed-1999-bitext}
I.~Dan Melamed. 1999.
\newblock \href {https://aclanthology.org/J99-1003} {Bitext maps and alignment
  via pattern recognition}.
\newblock \emph{Computational Linguistics}, 25(1):107--130.

\bibitem[{Mirjalili et~al.(2020)Mirjalili, Raschka, and Ross}]{semi-adv}
Vahid Mirjalili, Sebastian Raschka, and Arun Ross. 2020.
\newblock Privacynet: semi-adversarial networks for multi-attribute face
  privacy.
\newblock \emph{IEEE Transactions on Image Processing}, 29:9400--9412.

\bibitem[{Nicosia et~al.(2021)Nicosia, Qu, and Altun}]{taf-mt5}
Massimo Nicosia, Zhongdi Qu, and Yasemin Altun. 2021.
\newblock \href {https://doi.org/10.18653/v1/2021.findings-emnlp.279}
  {{T}ranslate {\&} {F}ill: {I}mproving zero-shot multilingual semantic parsing
  with synthetic data}.
\newblock In \emph{Findings of the Association for Computational Linguistics:
  EMNLP 2021}, pages 3272--3284, Punta Cana, Dominican Republic. Association
  for Computational Linguistics.

\bibitem[{Pires et~al.(2019)Pires, Schlinger, and Garrette}]{mbert}
Telmo Pires, Eva Schlinger, and Dan Garrette. 2019.
\newblock \href {https://doi.org/10.18653/v1/P19-1493} {How multilingual is
  multilingual {BERT}?}
\newblock In \emph{Proceedings of the 57th Annual Meeting of the Association
  for Computational Linguistics}, pages 4996--5001, Florence, Italy.
  Association for Computational Linguistics.

\bibitem[{Platt and Barr(1987)}]{constrained}
John Platt and Alan Barr. 1987.
\newblock Constrained differential optimization.
\newblock In \emph{Neural Information Processing Systems}.

\bibitem[{Ravfogel et~al.(2022)Ravfogel, Twiton, Goldberg, and
  Cotterell}]{rlace}
Shauli Ravfogel, Michael Twiton, Yoav Goldberg, and Ryan~D Cotterell. 2022.
\newblock \href {https://proceedings.mlr.press/v162/ravfogel22a.html} {Linear
  adversarial concept erasure}.
\newblock In \emph{Proceedings of the 39th International Conference on Machine
  Learning}, volume 162 of \emph{Proceedings of Machine Learning Research},
  pages 18400--18421. PMLR.

\bibitem[{Reid and Artetxe(2022)}]{reid2022role}
Machel Reid and Mikel Artetxe. 2022.
\newblock On the role of parallel data in cross-lingual transfer learning.
\newblock \emph{arXiv preprint arXiv:2212.10173}.

\bibitem[{Rongali et~al.(2020)Rongali, Soldaini, Monti, and
  Hamza}]{pointer-generator}
Subendhu Rongali, Luca Soldaini, Emilio Monti, and Wael Hamza. 2020.
\newblock Don’t parse, generate! a sequence to sequence architecture for
  task-oriented semantic parsing.
\newblock In \emph{Proceedings of The Web Conference 2020}, pages 2962--2968.

\bibitem[{Ruder et~al.(2021)Ruder, Constant, Botha, Siddhant, Firat, Fu, Liu,
  Hu, Garrette, Neubig, and Johnson}]{xtremer}
Sebastian Ruder, Noah Constant, Jan Botha, Aditya Siddhant, Orhan Firat, Jinlan
  Fu, Pengfei Liu, Junjie Hu, Dan Garrette, Graham Neubig, and Melvin Johnson.
  2021.
\newblock \href {https://doi.org/10.18653/v1/2021.emnlp-main.802}
  {{XTREME}-{R}: Towards more challenging and nuanced multilingual evaluation}.
\newblock In \emph{Proceedings of the 2021 Conference on Empirical Methods in
  Natural Language Processing}, pages 10215--10245, Online and Punta Cana,
  Dominican Republic. Association for Computational Linguistics.

\bibitem[{Sherborne and Lapata(2022)}]{prev-adv}
Tom Sherborne and Mirella Lapata. 2022.
\newblock \href {https://doi.org/10.18653/v1/2022.acl-long.285} {Zero-shot
  cross-lingual semantic parsing}.
\newblock In \emph{Proceedings of the 60th Annual Meeting of the Association
  for Computational Linguistics (Volume 1: Long Papers)}, pages 4134--4153,
  Dublin, Ireland. Association for Computational Linguistics.

\bibitem[{Wang et~al.(2020)Wang, Lipton, and Tsvetkov}]{negative-interference}
Zirui Wang, Zachary~C. Lipton, and Yulia Tsvetkov. 2020.
\newblock \href {https://doi.org/10.18653/v1/2020.emnlp-main.359} {On negative
  interference in multilingual models: Findings and a meta-learning treatment}.
\newblock In \emph{Proceedings of the 2020 Conference on Empirical Methods in
  Natural Language Processing (EMNLP)}, pages 4438--4450, Online. Association
  for Computational Linguistics.

\bibitem[{Winata et~al.(2021)Winata, Cahyawijaya, Liu, Lin, Madotto, and
  Fung}]{mmt-effective-cs}
Genta~Indra Winata, Samuel Cahyawijaya, Zihan Liu, Zhaojiang Lin, Andrea
  Madotto, and Pascale Fung. 2021.
\newblock \href {https://doi.org/10.18653/v1/2021.calcs-1.20} {Are multilingual
  models effective in code-switching?}
\newblock In \emph{Proceedings of the Fifth Workshop on Computational
  Approaches to Linguistic Code-Switching}, pages 142--153, Online. Association
  for Computational Linguistics.

\bibitem[{Wu and Dredze(2019)}]{freezing}
Shijie Wu and Mark Dredze. 2019.
\newblock \href {https://doi.org/10.18653/v1/D19-1077} {Beto, bentz, becas: The
  surprising cross-lingual effectiveness of {BERT}}.
\newblock In \emph{Proceedings of the 2019 Conference on Empirical Methods in
  Natural Language Processing and the 9th International Joint Conference on
  Natural Language Processing (EMNLP-IJCNLP)}, pages 833--844, Hong Kong,
  China. Association for Computational Linguistics.

\bibitem[{Xue et~al.(2022)Xue, Barua, Constant, Al-Rfou, Narang, Kale, Roberts,
  and Raffel}]{xue-etal-2022-byt5}
Linting Xue, Aditya Barua, Noah Constant, Rami Al-Rfou, Sharan Narang, Mihir
  Kale, Adam Roberts, and Colin Raffel. 2022.
\newblock \href {https://doi.org/10.1162/tacl_a_00461} {{B}y{T}5: Towards a
  token-free future with pre-trained byte-to-byte models}.
\newblock \emph{Transactions of the Association for Computational Linguistics},
  10:291--306.

\bibitem[{Xue et~al.(2021)Xue, Constant, Roberts, Kale, Al-Rfou, Siddhant,
  Barua, and Raffel}]{mt5-xue}
Linting Xue, Noah Constant, Adam Roberts, Mihir Kale, Rami Al-Rfou, Aditya
  Siddhant, Aditya Barua, and Colin Raffel. 2021.
\newblock \href {https://doi.org/10.18653/v1/2021.naacl-main.41} {m{T}5: A
  massively multilingual pre-trained text-to-text transformer}.
\newblock In \emph{Proceedings of the 2021 Conference of the North American
  Chapter of the Association for Computational Linguistics: Human Language
  Technologies}, pages 483--498, Online. Association for Computational
  Linguistics.

\bibitem[{Zhang et~al.(2018{\natexlab{a}})Zhang, Lemoine, and Mitchell}]{bias}
Brian~Hu Zhang, Blake Lemoine, and Margaret Mitchell. 2018{\natexlab{a}}.
\newblock \href {https://doi.org/10.1145/3278721.3278779} {Mitigating unwanted
  biases with adversarial learning}.
\newblock In \emph{Proceedings of the 2018 AAAI/ACM Conference on AI, Ethics,
  and Society}, AIES '18, page 335–340, New York, NY, USA. Association for
  Computing Machinery.

\bibitem[{Zhang et~al.(2018{\natexlab{b}})Zhang, Riesa, Gillick, Bakalov,
  Baldridge, and Weiss}]{langid}
Yuan Zhang, Jason Riesa, Daniel Gillick, Anton Bakalov, Jason Baldridge, and
  David Weiss. 2018{\natexlab{b}}.
\newblock \href {https://doi.org/10.18653/v1/D18-1030} {A fast, compact,
  accurate model for language identification of codemixed text}.
\newblock In \emph{Proceedings of the 2018 Conference on Empirical Methods in
  Natural Language Processing}, pages 328--337, Brussels, Belgium. Association
  for Computational Linguistics.

\end{thebibliography}
\bibliographystyle{acl_natbib}

\appendix

\begin{table*}[t]
\begin{center}
    \includegraphics[trim=3cm 0cm 3cm 1.85cm, clip,angle=90,width=0.9\paperwidth]{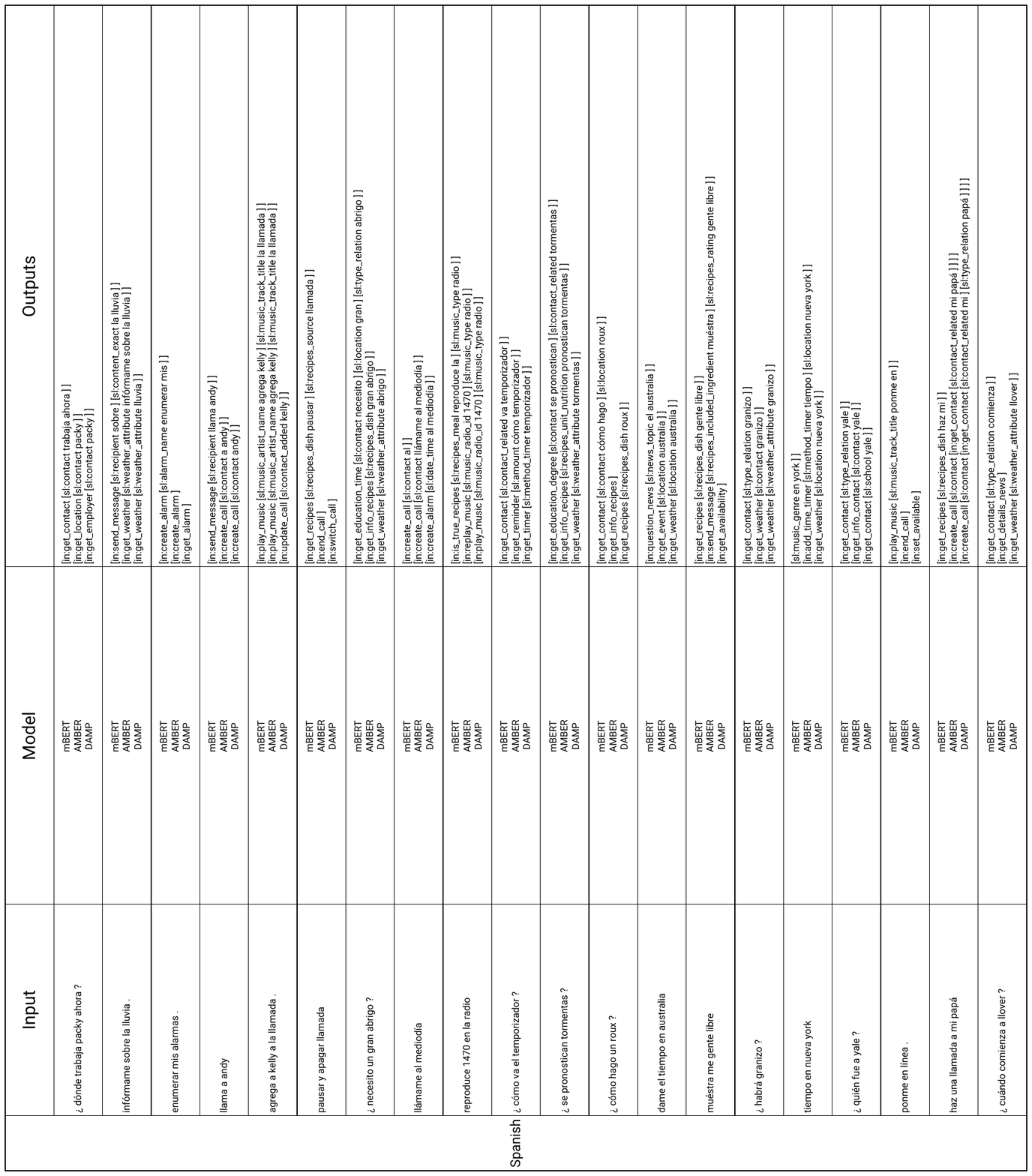}
    \caption{Full Table of 100 Sampled Spanish Results from Qualitative Analysis.}
    \label{first-qual}
\end{center}
\end{table*}
\begin{table*}[t]
\begin{center}
    \includegraphics[trim=3cm 0cm 3cm 1.85cm, clip,angle=90,width=0.9\paperwidth]{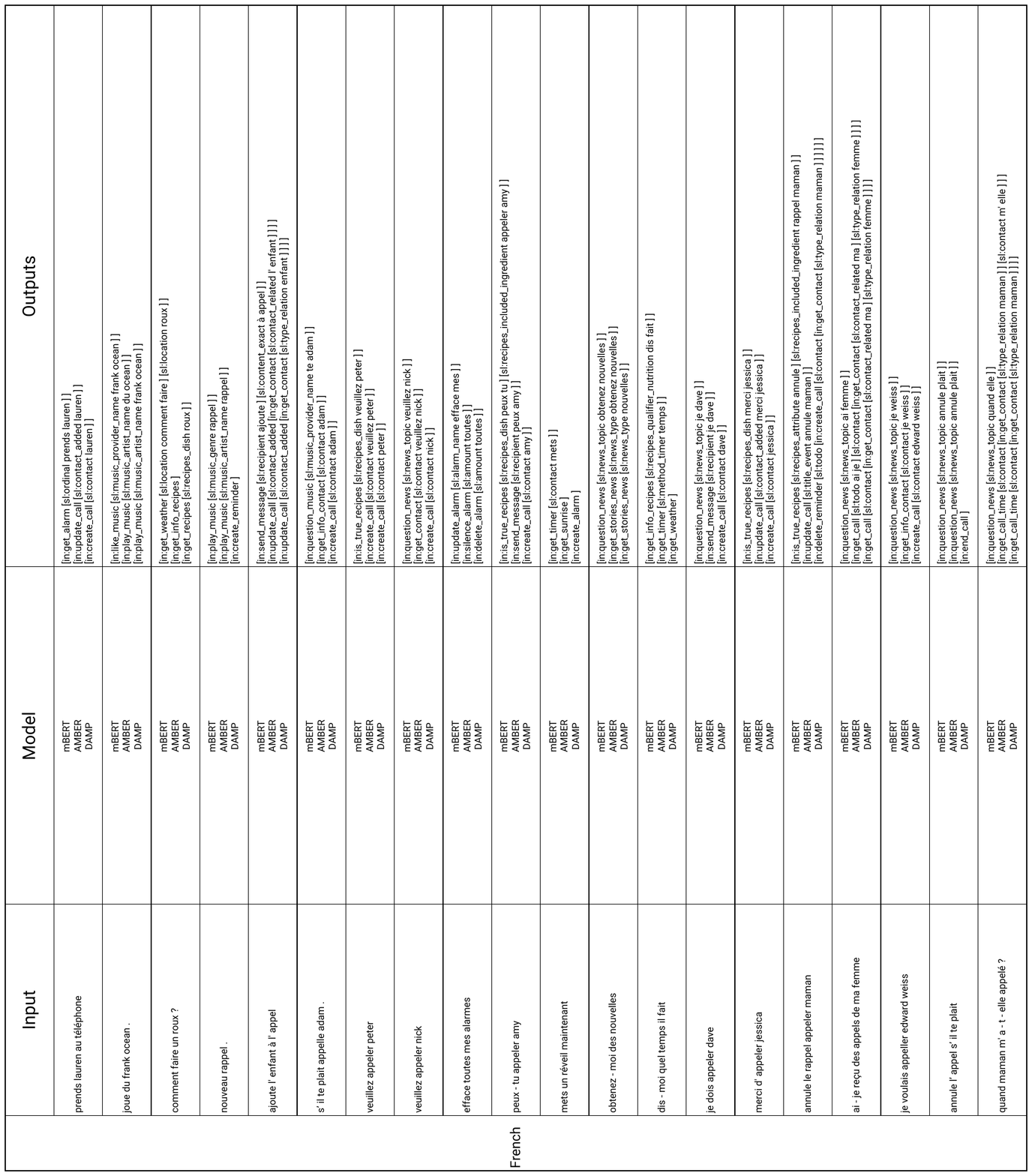}
    \caption{Full Table of 20 Sampled French Results from Qualitative Analysis.}
    \label{fr-qual}
\end{center}
\end{table*}
\begin{table*}[t]
\begin{center}
    \includegraphics[trim=3cm 0cm 3cm 1.85cm, clip,angle=90,width=0.9\paperwidth]{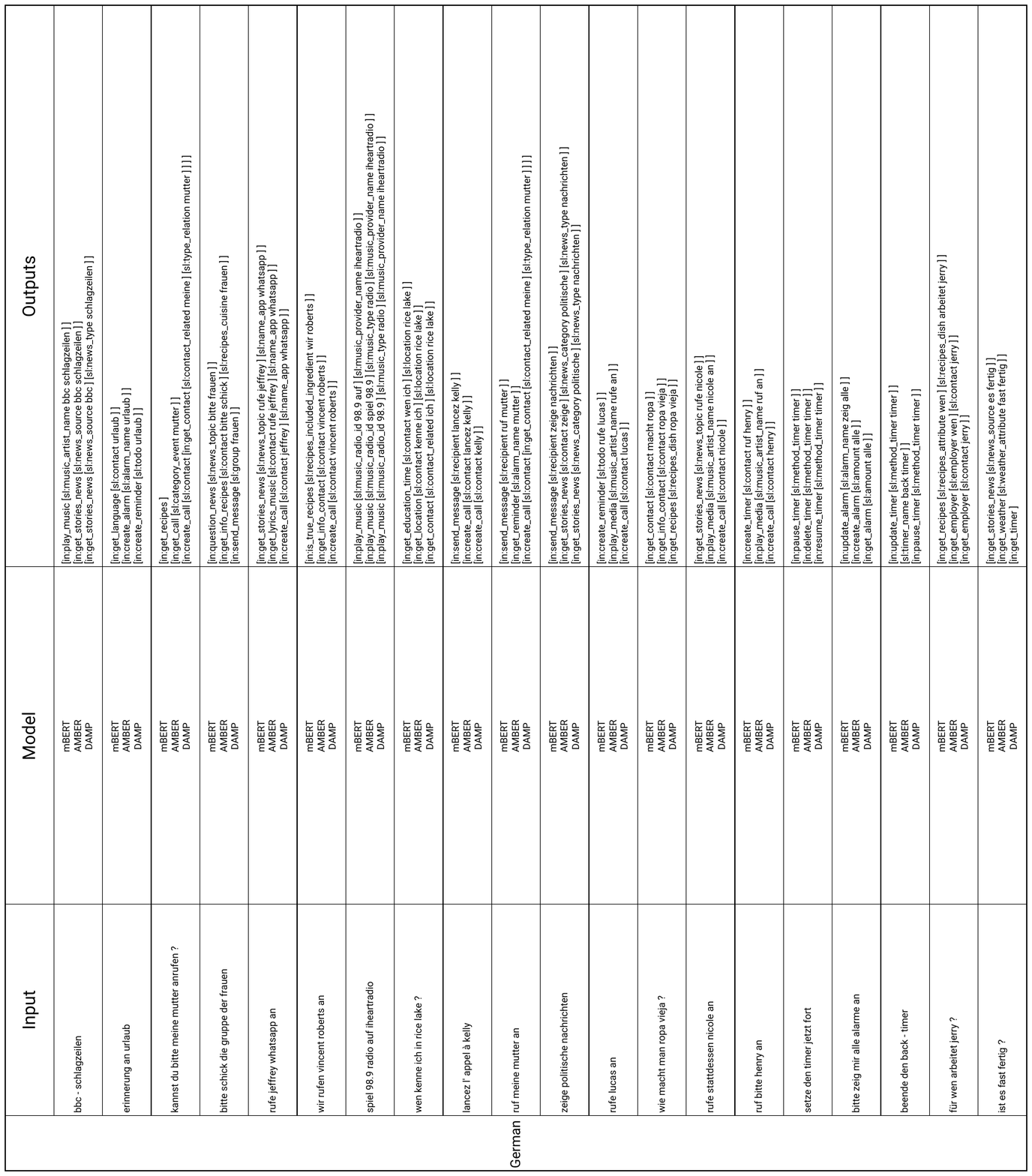}
    \caption{Full Table of 20 Sampled German Results from Qualitative Analysis.}
    \label{de-qual}
\end{center}
\end{table*}
\begin{table*}[t]
\begin{center}
    \includegraphics[trim=3cm 0cm 3cm 1.85cm, clip,angle=90,width=0.9\paperwidth]{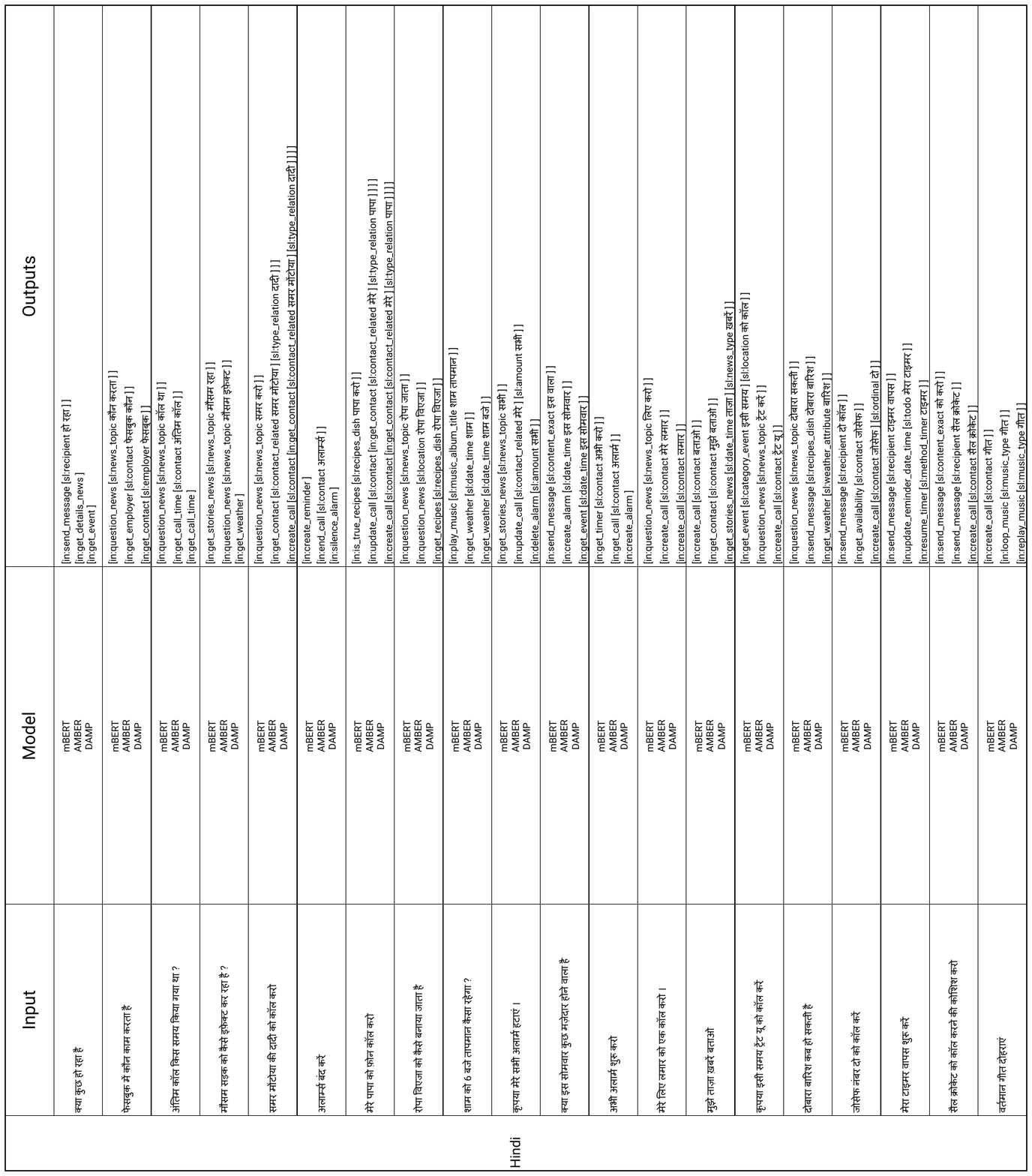}
    \caption{Full Table of 20 Sampled Hindi Results from Qualitative Analysis.}
    \label{th-qual}
\end{center}
\end{table*}
\begin{table*}[t]
\begin{center}
    \includegraphics[trim=3cm 0cm 3cm 1.85cm, clip,angle=90,width=0.9\paperwidth]{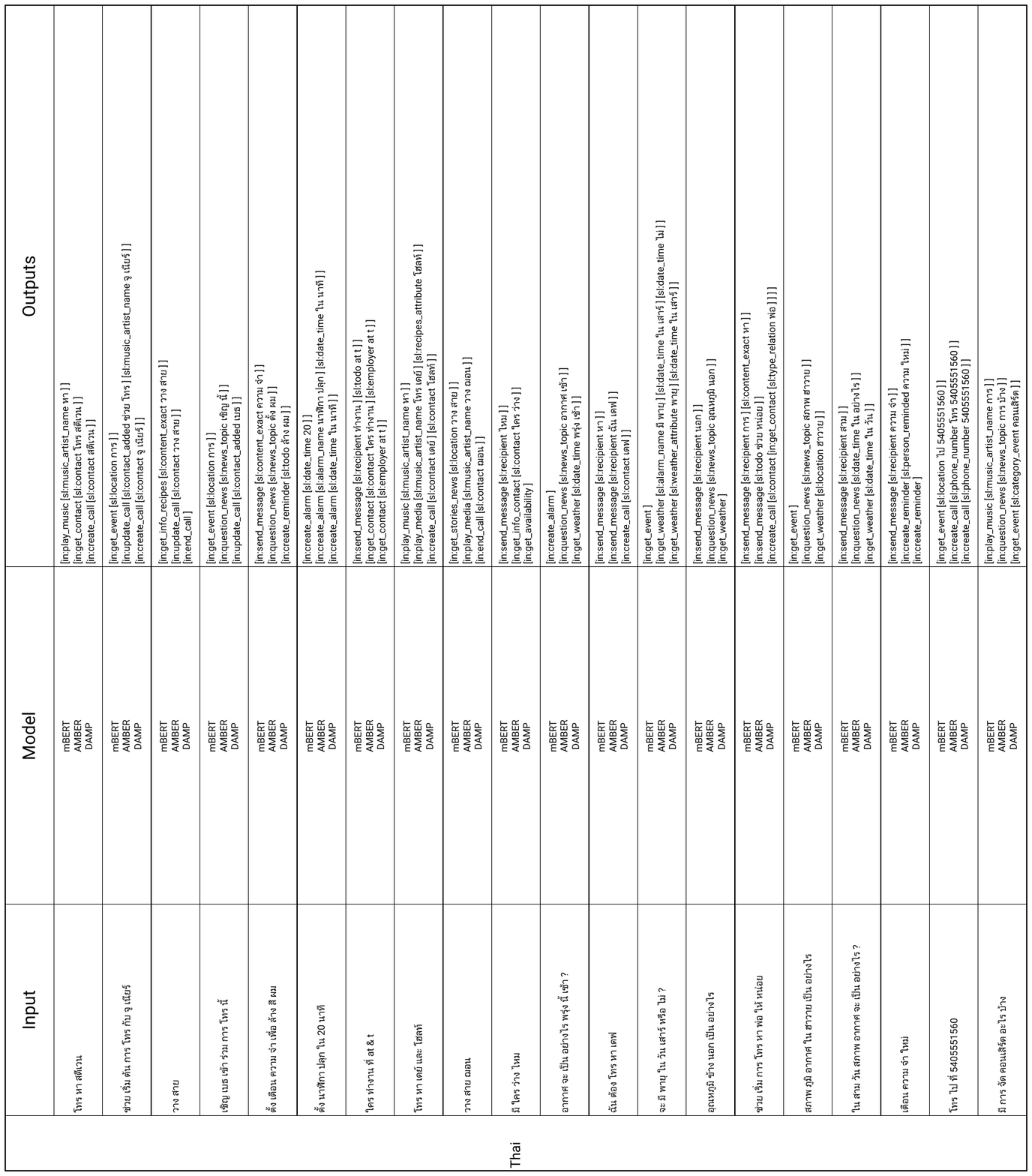}
    \caption{Full Table of 20 Sampled Thai Results from Qualitative Analysis.}
    \label{final-qual}
\end{center}
\end{table*}
\begin{table*}[t]
\begin{center}
\end{center}
\end{table*}
\end{document}